\pdfoutput=1

\documentclass[a4paper, 11pt]{article}

\usepackage[final]{acl}

\usepackage{times}
\usepackage{latexsym}

\usepackage[T1]{fontenc}

\usepackage[utf8]{inputenc}

\usepackage{microtype}

\usepackage{inconsolata}

\usepackage{graphicx} 
\usepackage{amsmath, amsfonts}
\usepackage{mathtools}
\usepackage{subcaption}
\usepackage{url}
\usepackage[export]{adjustbox}

\newcommand{\Rspace}{\mathbb{R}}
\newcommand{\VR}[2]{\mathrm{VR}_{#1}({#2})}
\newcommand{\PD}{\mathrm{PD}}
\newcommand{\abs}[1]{\left|{#1}\right|}
\newcommand{\norm}[1]{||\,{#1}\,||}
\newcommand{\card}[1]{|\,{#1}\,|}

\title{The Shape of Word Embeddings: Quantifying Non-Isometry With Topological Data Analysis}

\author{Ond\v{r}ej Draganov \\
  Institute of Science and Technology Austria (ISTA)\\
  Am Campus 1, 3400 \\
  Klosterneuburg, Austria \\
  \texttt{ondrej.draganov@ist.ac.at} \\\And
  Steven Skiena \\
  Dept. of Computer Science \\
  Stony Brook University \\
  Stony Brook, NY 11794-2424 USA \\
  \texttt{skiena@cs.stonybrook.edu} \\}

\begin{document}

\maketitle

\thispagestyle{plain}
\pagestyle{plain}

\begin{abstract}
    Word embeddings represent language vocabularies as clouds of $d$-dimensional points. We investigate how information is conveyed by the general shape of these clouds, instead of representing the semantic meaning of each token. Specifically, we use the notion of persistent homology from topological data analysis (TDA) to measure the distances between language pairs from the shape of their unlabeled embeddings. These distances quantify the degree of non-isometry of the embeddings. To distinguish whether these differences are random training errors or capture real information about the languages, we use the computed distance matrices to construct language phylogenetic trees over 81 Indo-European languages. Careful evaluation shows that our reconstructed trees exhibit strong and statistically-significant similarities to the reference.
\end{abstract}

\section{Introduction}

Does the {\em shape} of an unlabeled, monolingual word embedding carry relevant information about the language it represents?
Word embeddings are well-established objects of interest in natural language processing, being $d$-dimensional vector representations that capture the semantics of each vocabulary word.
The vocabulary of language $L$ can thus be viewed as a cloud of points, whose geometric and structural properties encode considerable information about the language.
In this paper, we demonstrate that, even after disassociated from their bindings to particular words, the ``shape'' of these point clouds correlates with the history of the languages they represent. We use techniques from {\em topological data analysis} (TDA), a field studying spatial aspects of data, to quantify this correlation.

How much shapes of word embeddings differ, and for what reasons, is an ongoing debate with important consequences. Many authors assume that the word embeddings of different languages are essentially isometric: meaning they only differ by an orthogonal transformation.
Such assumptions are necessary, for example, to construct translators between languages via bilingual lexicon induction \citep{barone-2016, Conneau-2018, zhang-2017}. This assumption has been disputed by \citet{sogaard2018-limitations} or \citet{patra2019}. Later \citet{vulic2020-good-isomorphic} argued that observed deviations from isometry are largely due to insufficient data or training in construction of embeddings for some languages. 
In this paper, we demonstrate that these observed non-isometries across languages contain real information and are thus not random training artifacts.
We analyze FastText embeddings of 81 Indo-European languages and quantify non-isometry among the 10k most frequent tokens of each language with TDA-based distance matrices. To show that those distances contain real information we construct phylogenetic trees and compare these to a gold-standard reference tree from Ethnologue \cite{ref:ethnologue}.

The history of language evolution is typically studied by identifying {\em cognate} word pairs of languages $L_1$ and $L_2$, namely homologous word forms inherited from the common ancestor of these languages.
Cognates usually share some level of surface-form similarity between them, with language pairs sharing more cognates presumed to be more closely related.
The identification of cognate pairs traditionally requires extensive labor by skilled linguists, although computational methods for detecting cognate pairs are emerging \cite{kondrak-2001-identifying,lefever-etal-2020-identifying,st-arnaud-etal-2017-identifying}.

An interesting aspect of TDA on word embeddings is that it makes no assumption or use of translation pairs, surface similarities between words, or cognate pairs.
Word embeddings are treated as {\em unlabeled} sets of points, shorn of any binding to corresponding tokens.
TDA compares the relative {\em shape} of word embeddings in terms of structural properties like components of connectivity and holes.
That we demonstrate even a weak ability to reconstruct language phylogenies through TDA is quite surprising, and suggests that there are hitherto unknown structures reflected in word embeddings that go beyond the semantics of individual words.

Our major contributions in this paper include:

    \paragraph{Further evidence that even within the Indo-European family the non-isometry between unlabeled word embeddings is not completely erroneous.} The fact that we reconstruct phylogenetic trees that, despite not very precise, are considerably better than chance, suggests that there is some real information represented in the differences between unlabeled shapes of embeddings. It is fair to argue that more careful training, normalization and lemmatisation could yield more similar embeddings that miss the information; were that the case, it would suggest that the differences mitigated by those processes correlate with language history, which is an equally interesting observation.

    \paragraph{A case study of natural generalizations of previously used methods to quantify non-isometry.} TDA provides a relatively new set of tools for data science able to capture features often missed by more traditional approaches.
    Due to the general framework it provides, there are many parameters to choose from and a priori no obvious canonical choices.    
    Our analysis on word embeddings represents a rigorous evaluation for an interesting use-case of this technology.
    
    A particular instance of TDA was used in \citet{patra2019, vulic2020-good-isomorphic} as an approximation \cite{chazal2009-gromov-hausdorff} of difficult-to-compute Gromov-Hausdorff distance to quantify non-isometry of word embeddings. In the TDA terminology, they computed \emph{bottleneck distance between degree~$0$ persistent diagrams}---we additionally explore degrees $1$ and $2$, and three different distances in addition to bottleneck; each combination for both Euclidean and cosine metric on the word embedding. Altogether this totals to 24 different combinations of parameters, and for each we obtain a distance matrix between the studied languages. All those notions of distances are invariant under orthogonal transformations\footnote{Cosine distances can change with translation---a shift of the whole point cloud.}, and therefore serve as quantifications for non-isometry. The alternative combinations can be better suited for studying word embeddings, e.g., by being less prone to irrelevant outliers that can artificially significantly increase the GH distance.
    Our analysis also shows that degree~$2$, which is not that often used in large high-dimensional data because of the expense in constructing them and difficulty to interpret them, yield statistically-strong results in our experiments, in some cases outperforming degrees~0 and~1.

    \paragraph{Statistical Evaluation of TDA-Based Language Distances with Phylogenetic Trees.}Evaluating the information content of language distance matrices inferred using TDA is best done by converting these matrices into phylogenetic trees, and then assessing the quality of these trees against a reference standard.
    Each of 24 different TDA matrices, constructed based on different combinations of parameters, was evaluated on two popular tree construction algorithms---UPGMA and neighbor joining---and compared against a gold-standard language tree from Ethnologue under six different tree-similarity metrics.

    Certain choices of parameters prove less capable of recognizing the structural similarities inherent in word embeddings.
    But permutation tests show that for 484 out of 864 combinations of parameters\footnote{See Section~\ref{sec:methods} for a break-down of the $864=2\cdot 3\cdot 4\cdot 3\cdot 2\cdot 6$ combinations of parameters.}, the quality of the reconstructed trees were significant to the 0.05-level, and many substantially stronger than that.
    Even stronger bounds come in the number of standard deviations our TDA trees are closer to the gold-standard tree compared to the mean statistical background.
    Many of our TDA-based trees sit from four to seven $\sigma$ from the mean.
    Under assumptions of normality, the maximum achieved $6.87 \sigma$ corresponds to a Bonferroni-corrected $p$-value of $2.77 \times 10^{-9}$ of seeing a result like this by chance.

\paragraph{} We presume a reader from computational linguistics, with a basic familiarity with the concepts and literature of word embeddings and language phylogenies, but no prior exposure to topological data analysis.
Our paper is organized as follows.
Important TDA concepts like persistence diagrams are reviewed in Section~\ref{sec:tda}.
Algorithms for constructing language phylogenies from distance matrices and evaluating the resulting trees are presented in Section~\ref{sec:language_phylogenies}.
We describe our TDA-based analysis pipeline in Section~\ref{sec:methods}
with computational results reported in Section~\ref{sec:experiments}, and conclude with Section~\ref{sec:conclusions}.

Our primary goal in this work is to initiate the study of the shape of languages via word embeddings, {\em not} to advance the state-of-the-art in language phylogeny reconstruction.
Indeed, cognate-based analysis should clearly dominate our unlabeled point methods except perhaps in pathological situations, such as completely undeciphered written languages.
But we do believe that the new TDA-based tools hold promise for richer computational language analysis, and raise interesting questions about which properties in language-space the topological features we employ correspond to.

\section{Topological Data Analysis}\label{sec:tda}

Topological data analysis (TDA) is a growing field that applies methods developed for studying shapes to data, both geometrical and abstract, to extract features that are often not captured by classical approaches of data science. In the case of points or vectors in $2$D or $3$D, we observe patterns like clusters of points, loops, voids or tunnels.

In this paper, the input is a word embedding of a language---collection of points in high-dimensional Euclidean space, one for each token---and the output is a summary of its topological features. We compute such summaries for embeddings of different languages, and use them to compare the structural similarity. As orthogonal transformations would preserve the summaries, those differences quantify non-ismoetry of the different embeddings.

To keep the explanation intuitive, we stay with low dimensional point sets, but the general framework is limited neither by dimension of the data nor by the Euclidean metric---indeed, we use $300$-dimensional word embedddings and impose both Euclidean and cosine metrics.
For a gentle introduction to TDA see \citet{short_course_in_computational_topology_book}.
For more technical details, see \citet{computational_topology_book}. For applications of TDA to a wide variety of real-world data, see the database of non-theoretical uses of topology \citep{DONUT}.

\subsection{Persistent Homology}\label{sec:persistent_homology}

\begin{figure*}[tbh]
    \centering
    \includegraphics[width=.31\textwidth]{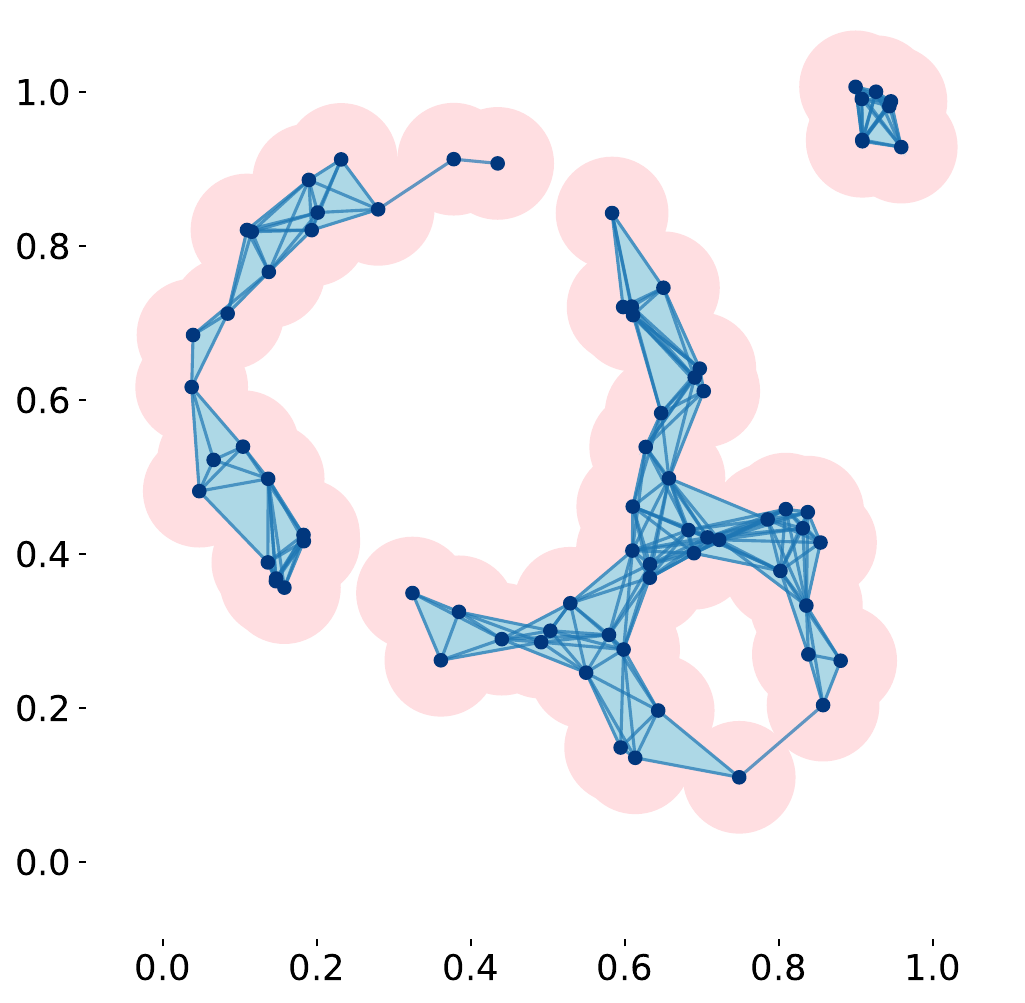}
    \hspace{3mm}%
    \includegraphics[width=.31\textwidth]{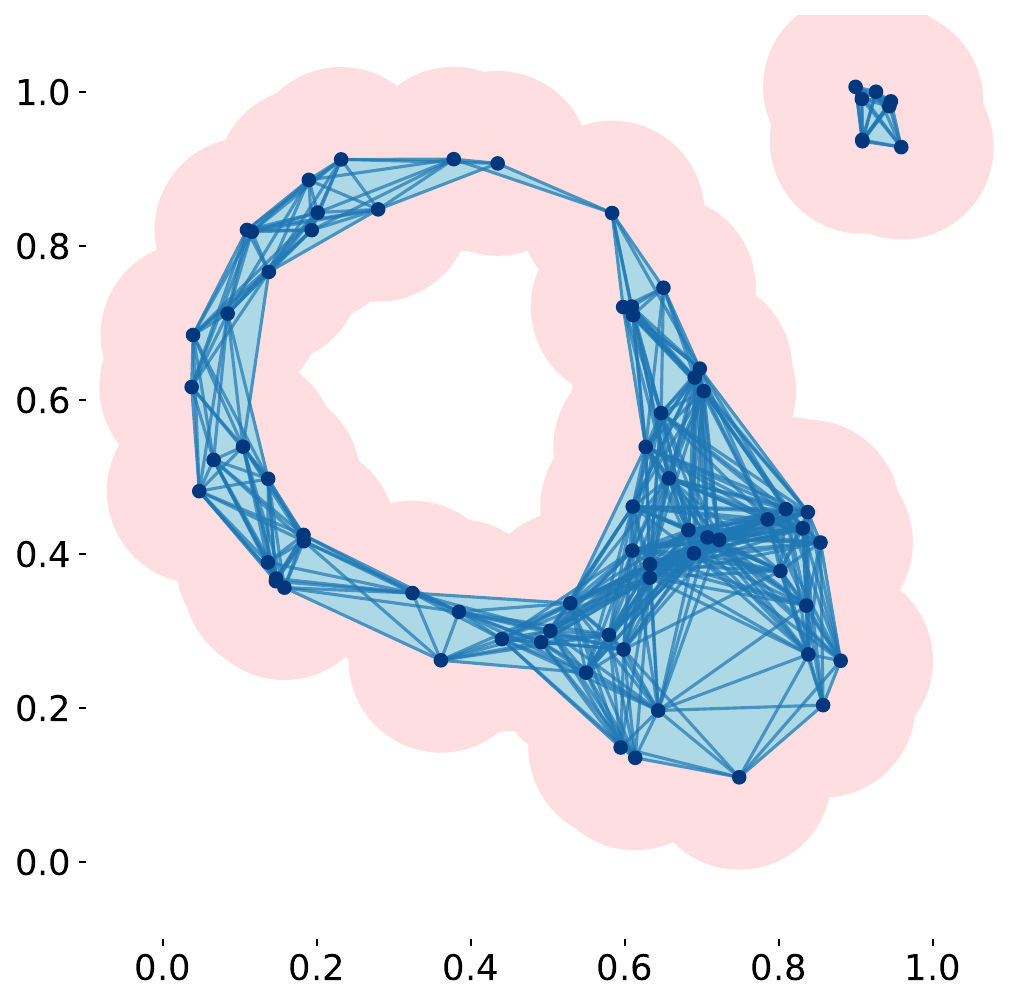}
    \hspace{3mm}%
    \includegraphics[width=.31\textwidth]{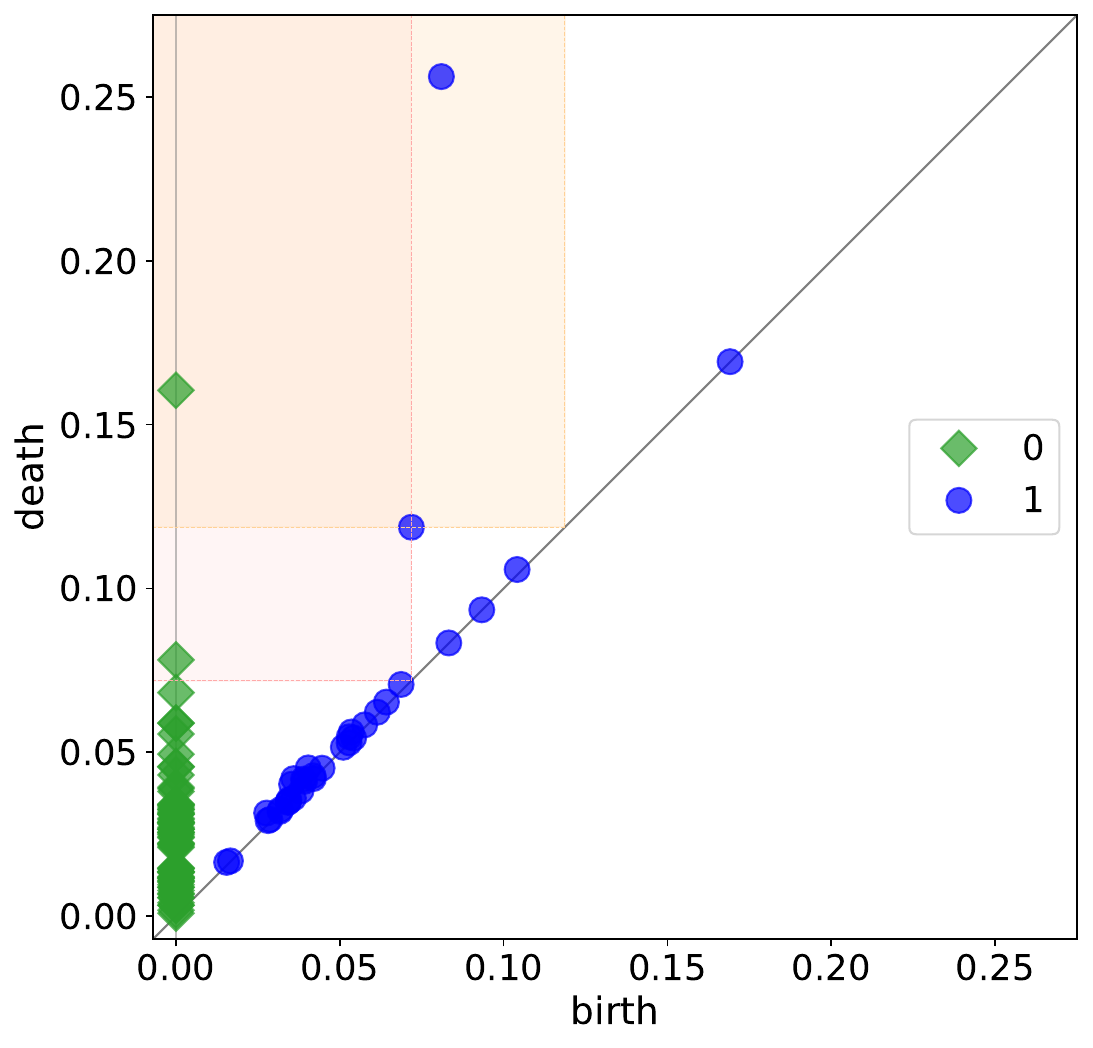}
    \caption{Unions of disks centered at data points in a Euclidean plane define a `shape' that varies with the radius of the disks. In our setting, each point is one token of a word embedding, and the space is $\Rspace^{300}$ rather than~$\Rspace^2$. Persistent homology studies different types of `gaps' within this growing shape---$0$-dimensional gaps between components of connectivity, and $1$-dimensional holes within the shape. Each such feature is born at some radius, and dies at some later radius, spanning an interval $[r_b, r_d)$. Those intervals are summarised in a persistence diagram; each interval is represented by a dot with coordinates $(r_b, r_d)$.
    The \emph{left} and \emph{middle} figures correspond to birth and death radii of the loop in the lower-right corner: $[r_b, r_d)=[0.072, 0.119)$. Overlaid is the Vietoris-Rips complex discretizing the pink shape to allow for computations (see \ref{sec:vietoris-rips}). On the \emph{right} is the persistence diagram of the growing disks with highlighted blocks corresponding to radii $r_b$, $r_d$.
    }
    \label{fig:growing_disks}
\end{figure*}

We describe the idea of persistent homology in a simple intuitive setting---a data set of points in a Euclidean plane. To define the `shape' of such a set, choose a radius, place a disk of that radius centered at each point, and consider the union of the disks; see Figure~\ref{fig:growing_disks}. Two simple descriptors for such a shape are the number of connected components and the number of holes---\emph{Betti numbers}\footnote{
The formal definition of Betti numbers depends on the choice of coefficients for homology. We use the field of size two, which is the standard choice for persistent homology.
}
$\beta_0$ and $\beta_1$, respectively. The notion is formally defined by \emph{homology} from the mathematical field of algebraic topology; see, e.g., \citet{hatcher_book}.

For different radii of the disks, we get different Betti numbers. Consider growing the radius from zero to infinity. We can plot the Betti numbers depending on the radius, and treat the curves, $\beta_0(r)$, $\beta_1(r)$, as descriptors for the shape of our point set. The problem is that such a curve can differ a lot when we slightly perturb the points. For a stable descriptor, we turn to \emph{persistent homology}, which, in addition, pairs the \emph{birth} and \emph{death} radii of each component or hole. Components are all born at radius 0 and each time two components merge at radius $r$, we get a birth-death pair $[0,r)$. For a hole or a loop, there is a minimum birth radius, $r_b$, for which the disks enclose a space in the plane, separating it from its surroundings, and there is a larger minimum death radius, $r_d$, for which the area we previously enclosed is fully covered by the disks. The interval $\left[r_b, r_d\right)$, often called a \emph{bar}, is a single feature recorded by persistent homology. It is the interval of radii during which this particular hole contributes ``plus one'' to the corresponding Betti number.

We collect all such intervals. Each interval is described by its two endpoints, so we can plot them all in a scatter plot with axes ``birth radius'' and ``death radius''; see Figure~\ref{fig:growing_disks}(c). This plot is the \emph{persistence diagram} of the data. For a point set in a plane, we get two persistence diagrams: one for components of connectivity---degree 0---and one for `holes' or `loops'---degree 1. Usually we plot both in the same figure, separating them by color or shape of the markers.

When we increase the dimension of the data, we can have more different kinds of features.
In 3D, we grow balls rather than disks around the points. A degree~2 feature is an uncovered space completely surrounded by covered space---like a hole completely inside a Swiss cheese. A degree~1 feature is a loop drawn into the union of the balls that cannot be contracted to a point within that union. As loops in 3D, a $2$-dimensional feature in higher dimensional data would not be a completely enclosed space, but rather a drawing of a $2$-dimensional sphere that cannot be contracted to a point.

Although our actual data is $300$-dimensional, we only consider features of degree~$0$, $1$ and~$2$, mainly for computational reasons.
Examples of diagrams from our analysis are in Appendix Figure~\ref{fig:persistence_diagrams}.

Persistent homology is stable against perturbations. That is, if the data points are perturbed a bit, the points on the persistence diagrams are also only perturbed a bit \cite{CoEdHa07}.
Note that in contrast, it is not very stable against outliers, which is a limitation of the tool that can be somewhat mitigated with the right choice of distance between persistence diagrams.

\subsection{Distances Between Persistence Diagrams}\label{sec:distances_between_diagrams}

In this section, we describe several established notions of distances between persistent diagrams. The list we use is by no means exhaustive---see, e.g., \citet{vectorization_survey2023} for a survey of methods, and \citet{perslay2020} for an approach unifying many of those methods under a common framework. We chose four distances that are widely used in theory or in applications and have readily available implementations. Two of those are defined directly on the diagrams, while the other two first vectorize each diagram and then compare the vectors with the standard Euclidean metric.
Note two properties that make defining distances somewhat involved: firstly, different persistence diagrams have generally different number of points with no canonical matching between them; secondly, slightly perturbing data can create or remove points at the diagonal, which needs to be taken into consideration.

We fix two persistent diagrams, $\PD_1$ and $\PD_2$, describing topological features of the same degree. An element $r$ of $\PD_i$ is a point in a plane representing an interval, and we will denote its endpoints by $r=[r_b, r_d)$ for ``birth'' and ``death''.

\paragraph{Bottleneck distance.} The first distance that was proved to be stable \cite{CoEdHa07} is the \emph{bottleneck distance}, which is defined as ``the biggest difference for the best matching''. A \emph{matching}, $\mathcal{M}$, between two persistence diagrams is a pairing of the points they contain, with an added flexibility---we can match points with any points on the diagonal. That is, if $(r,s)\in \mathcal{M}$, then either $r\in\PD_1$ and $s\in\PD_2$ or only one of those holds and the other element is $[x,x)$ for some number $x$. The bottleneck distance between $\PD_1$ and $\PD_2$ is then
\begin{multline*}
    W_\infty(\PD_1, \PD_2) = \\ \inf_{\mathcal{M}} \max_{(r, s)\in\mathcal{M}} \max\{\abs{r_b-s_b}, \abs{r_d-s_d}\}.
\end{multline*}
Bottleneck distance is a classical notion, but it is computationally costly since we have to search the space of matchings.
It is comparing the similarity of the diagrams ``on the nose''. This is well-suited for those applications in which objects are considered similar only if their ``hole-structure'' is almost identical, but can be too penalising when we are interested more vaguely in the distribution of different kinds of holes. It is also very sensitive to outliers.

\paragraph{(Sliced) Wasserstein distance.} The maxima in the bottleneck distance can be seen as $l_\infty$-norm, which suggests that they can be replaced by $l_p$ norms, for various $p\in\mathbb{N}$, to get different notions of distances. In particular, replacing both maxima by $l_1$-norm, we get \emph{$1$-Wasserstein distance} between persistence diagrams:\[
    W_1(\PD_1, \PD_2) = \inf_{\mathcal{M}} \sum_{(r,s)\in\mathcal{M}} \abs{r_b - s_b} + \abs{r_d - s_d}
\]
This distance takes each interval in either diagram into consideration, not just the worst match as the bottleneck distance. For discussion on stability with respect to data perturbations see \citet{CoEdHaMi10, SkTu23}. In our analysis, we use a computationally more viable variant---\emph{sliced Wasserstein distance} which approximates $1$-Wasserstein distance \cite{CaCuOu17}.

\paragraph{Persistence image.} Rather than directly comparing the diagrams, we can first transform each into a vector and then compare the vectors. One approach to vectorize a diagram is to blur the points and then treat the plot as a raster image, called \emph{persistence image} \cite{Ad17}. Persistence images are stable and approximate the $1$-Wasserstein distance.
More formally, we put a two-dimensional Gaussian over each dot in the persistence diagram, and consider their sum. Then we overlay a grid over the domain, and compute the integral of the sum of Gaussians over each grid square. This yields a value for a pixel. Since the dots further away from the diagonal are more relevant, there is a weight function defining the height of the Gaussian based on the distance away from the diagonal. The distance between two diagrams is then the Euclidean distance between their persistence images viewed as vectors.

Persistence images are easy to compute, but there is several parameters we need to fix---the size of the pixels and the grid, the $\sigma$-parameter of the Gaussians, and the weight function controlling the heights of the Gaussians based on the distance from the diagonal.

\paragraph{Bars statistics.} A somewhat naive, but surprisingly successful vectorization is to collect several simple statistics of the collection of dots (also called bars) in the persistence diagram \cite{pun2022, cang2015, vectorization_survey2023}. We used 40 numbers---10 statistics for 4 values. For each dot, $(b,d)$, the 4 values are birth, death, persistence and midpoint, i.e., $b$, $d$, $d-b$ and $\frac{d+b}{2}$, respectively. The 10 statistics are: median, standard deviation, interquartile range, full range, 10th percentile, 25th percentile, 75th percentile, 90th percentile, and entropy. To compare two diagrams, we compute the Euclidean distance between the vectorizations.

This vectorization method misses theoretical guarantees like the stability, but is very easy to compute, the meaning of the features is clear, and may mitigate the negative effects of outliers.

\section{Constructing Phylogenetic Trees}
\label{sec:language_phylogenies}

To evaluate whether the distances between languages calculated based on persistence homology are relevant, we use this data to attempt to reconstruct the historical evolution of Indo-European languages. We perform a hierarchical clustering, obtain a phylogenetic tree, and compare this tree to a ground-truth reference tree.

We emphasize that reconstructing language phylogenies is not the goal of our analysis, but rather we use it as a tool to evaluate whether the computed distances carry a relevant signal, meaning the unlabeled topological ``shape'' of word embeddings retain and reflect broader properties of the underlying languages. Phylogeny reconstruction is an enormous area within computational linguistics, and space does not permit us to make a substantial survey of work in this field.
See \citet{dunn2015language,pereltsvaig2015indo,pompei2011accuracy} for recent reviews of this literature.

In this paper, we experiment with two popular agglomerative hierarchical clustering algorithms for reconstructing phylogenetic trees from a pairwise distance matrix, $d(\cdot, \cdot)$.
We start with each language as an individual cluster, and then we connect two closest clusters in each step. The difference is what ``closest'' means for each of the algorithms:

\paragraph{Unweighted Pair Group Method with Arithmetic Mean (UPGMA).} The UPGMA method \cite{SneathSokal1973_upgma} approximates distances between clusters as the average distance between the elements: for clusters $A$, $B$, the distance is \[
    d(A, B) \coloneqq \frac{1}{\card{A} \cdot \card{B}} \sum_{\substack{a\in A \\ b\in B}} d(a, b).
\]
Starting with singleton clusters, in each step it merges the two closest clusters into one.

\paragraph{Neighbor Joining (NJ).} The NJ method \cite{SaitouNei1987_nj} defines distances between clusters, $D(\cdot, \cdot)$, inductively through a merging process.
Let $C$ be the collection of current clusters. For each cluster, $a\in C$, define \[ \alpha(a)\coloneqq \frac{1}{\card{C}-2}\sum_{\substack{b\in C \\ b\neq a}} D(a,b).
\]
In each step, find a pair $a, b\in C$ minimizing $D(a,b) - \alpha(a) - \alpha(b)$. Then replace $a, b$ by the union, $a\cup b$, and define the new distances $D(a\cup b, c) \coloneqq (D(a,c) + D(b,c) - D(a,b))/2$ for each other $c$. The length of the new tree edge, $(a\cup b, a)$ is defined as $D(a,b) + \alpha(a) - \alpha(b)$, and symmetrically for $(a\cup b, b)$.

\section{Analysis Pipeline}\label{sec:methods}

This section details the data, the process of going from the data to trees, and finally the evaluation of the reconstructed trees. Our code for the analysis is published on GitHub.\footnote{\url{https://github.com/OnDraganov/shape-of-word-embeddings}}

\subsection{Data} \label{sec:data}

Our study requires two types of data over a large set of languages: (a) a source of directly comparable, high quality word embeddings, and (b) a ground-truth reference phylogenetic tree reflecting the origin and relative similarity of languages in this set.
We limit our attention here to the broad family of Indo-European languages for this study.

The historical origin of languages has been extensively studied, and much is known, but debate remains vigorous even for the Indo-European languages we study here \cite{gray2003language,pereltsvaig2015indo,longobardi2013toward}.
The reference tree we use comes from Ethnologue\footnote{\url{https://www.ethnologue.com}}, which is the most widely consulted inventory of the world's languages \cite{ref:ethnologue}. 
First published in 1951 and now in its 26th edition, Ethnologue currently records data about 7,168 living languages.
Although there is no universal agreement on language origins (see e.g. \citet{hammarstrom2015ethnologue} for a critique of Ethnologue), it provides a broadly acceptable reference tree for our evaluation purposes, particularly for the major Indo-European languages we consider here.
One downside of this reference is the lack of weights on the edges of the tree---we only get the topology. An alternative for a follow-up research could be to compare to other computationally reconstructed trees that are considered reliable, e.g., \citet{Serva_2008-Levenshtein}.

FastText\footnote{\url{https://fasttext.cc/}} \cite{GrBoGuJoMi18} provides a set of pre-trained, 300-dimensional word embeddings for 157 languages, trained on Common Crawl and Wikipedia.
There are 81 Indo-European languages that appear in the Ethnologue reference tree and also have pre-computed FastText embeddings.
To protect against the risk that low-resource languages might have less reliable embeddings, we will also evaluate filtrations to the most popular 30 and 50 languages. The Ethnologue reference trees for all filtrations are provided for inspection in Appendix Figure~\ref{fig:ethnologue_tree}.

\subsection{From Language Embeddings to a Tree} \label{sec:methods:pipeline}

In this section we describe our pipeline to go from the embeddings of languages to a phylogenetic tree, list the parameter choices at each step, and provide details about the implementation.

We start with a set of embeddings of languages. For each language, we compute a distance matrix between its words. For each such matrix, we use persistent homology and get a set of persistent diagrams. Computing distances between the persistent diagrams yields a distance matrix labeled by languages. From this language distance matrix we finally construct a phylogenetic tree.

\paragraph{Language embeddings.} As described in Section~\ref{sec:data}, we work with language embeddings from FastText.
We filter the data in two ways. Firstly, we use the $V=10,\!000$ most frequent tokens of each language---a threshold chosen to be as high as possible considering computational feasibility\footnote{Degree~2 PH is expensive; for $0$ and $1$, $V$ can be increased.}. Secondly, we only choose a subset of the languages to work with.
Because low-resource languages may skew our analysis, we filter the languages by the number of Wikipedia articles\footnote{\url{https://meta.wikimedia.org/wiki/List_of_Wikipedias}, extracted on November~8, 2023}, and consider the first 30, 50 or all 81 available Indo-European languages.

\paragraph{Token-to-token distance matrices.} As input to the persistent homology pipeline we use, for each language, we generate a $V \times V$ matrix, each entry being the distance between two 300-dimensional vectors, $u$, $v$. We use two different notions of distances: either the standard Euclidean distance, $\norm{u - v}$, or the cosine distance, $1 - a\cdot b \,/ \left( \norm{a} \cdot \norm{b} \right)$, which is often used in language processing.

\paragraph{Persistence diagrams.} For the token-to-token distance matrix of each language, we compute the persistence diagram\footnote{We use Vietoris-Rips complex, described in Section~\ref{sec:vietoris-rips}.} as described in Section~\ref{sec:tda}, using an efficient implementation of this computation, \texttt{Ripser}\footnote{\url{https://github.com/Ripser/ripser}} \citep{Bauer2021Ripser}. In this step, we choose the degree of the topological features to use---either 0, 1 or 2.

\paragraph{Languages distance matrix.} From the set of persistence diagrams, we get a single distance matrix representing the proximity of language pairs. The distance between diagrams is one of the four described in Section~\ref{sec:distances_between_diagrams}---this is the fourth parameter.

We use bottleneck distance implementation from \texttt{GUDHI} library\footnote{\url{https://gudhi.inria.fr}} \citep{gudhi:BottleneckDistance}, and slice Wasserstein and persistence image implementations from \texttt{persim} Python library\footnote{\url{https://github.com/scikit-tda/persim}}. Bar statistics are computed directly.

Persistence images take several parameters. Our choices depend on the embedding metric used and persistence diagram degree considered. For degrees $1$ and $2$ we use a $10\times 10$ grid. For degree $0$ it is just $1\times 10$ grid. The range of radii covered by the grid is $[0, 10]$ for Euclidean embedding metric, and $[0, 1]$ for cosine. Note that if there is a dot outside of this range in the persistence diagram, the Gaussian is still put over it, and spills into the considered range, so even those dots are still considered in the distance. The weight function is linear in all cases, $w(d-b) = d-b$. Careful optimization of these parameters might lead to better results.

For the bars statistics, in degree $0$ we only use the death, $d$, value, as $b=0$ for all dots. That is, the vectorization is 10-dimensional in this case rather than 40-dimensional as in the other cases.

\paragraph{Tree construction.} Finally, we construct a phylogenetic tree based on the language distance matrix. We try two different approaches. We use either of the two agglomerative clustering algorithms described in Section~\ref{sec:language_phylogenies}---neighbor joining (NJ) or unweighted pair group method with arithmetic mean (UPGMA). Both of those algorithms are implemented in \texttt{biopython} package\footnote{\url{https://biopython.org/wiki/Phylo}} under \texttt{DistanceTreeConstructor} class. Figure~\ref{fig:upgma_nj_trees} in Appendix gives examples of reconstructed and reference trees.

Two aspects of this approach make comparisons of thus reconstructed trees to the reference difficult. Firstly, the reference Ethnologue tree has no weights on edges, so we cannot meaningfully use the weights of the reconstructed trees either. Secondly, the Ethnologue tree is flat and neither binary nor rooted, while the reconstructed trees are generally somewhat deep binary trees. Our solution to this difference in topology is to test the reconstructed trees against their copies with permuted leaf labels.

We also conducted a second set of experiments to allay the concerns about the difference in height and topology between reconstructed and reference trees. We fix the topology of the reference Ethnologue tree, and try to assign languages to the leaves---to do that, we optimise the similarity of the path-distance matrix given by the labeled tree to the distance matrix computed using TDA. We then evaluate this optimized labeling against 100,000 random labelings. Consistent with the results reported below, we perform substantially better than chance for almost all combinations of parameters, typically by two or more standard deviations from the mean.
Details of this alternative analysis are presented in the Appendix~(\ref{sec:preserving-ethnologue-topology}).

\paragraph{Summary of the parameters.} To construct one tree we choose a number of languages (30, 50, 81), an embedding (Euclidean, cosine), a degree of persistence diagrams (0, 1, 2), a distance between persistence diagrams (bottleneck, sliced Wasserstein, persistence image, bars statistics), and a construction of a tree (UPGMA, NJ)---altogether $3\cdot 2\cdot 3\cdot 4\cdot 2 = 144$ variants. For each there are six tree distances to evaluate how well we reconstructed the phylogeny, as described below.

\begin{figure*}[ht]
    \centering
    \includegraphics[width=0.93\linewidth]{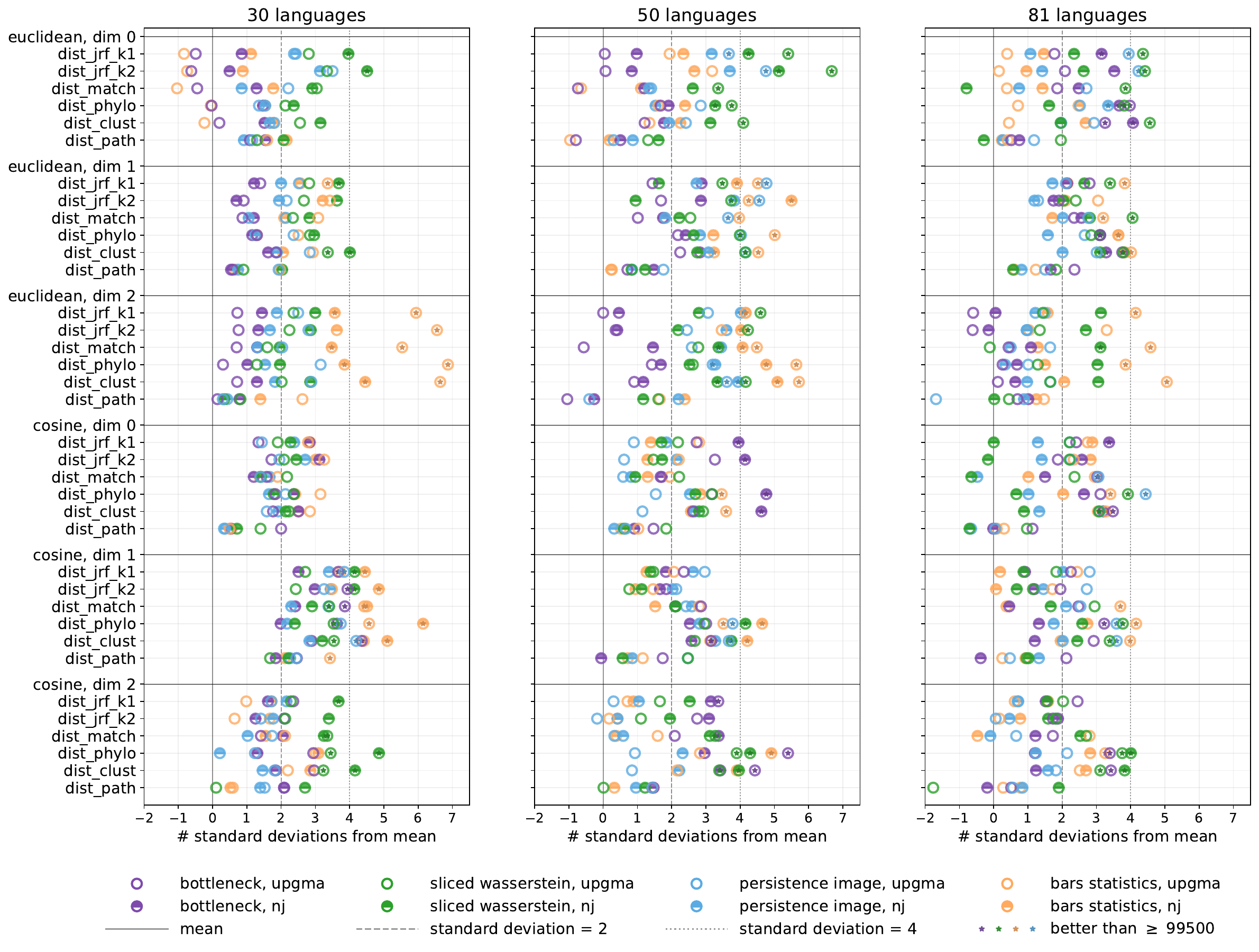}
    \caption{The statistical significance of TDA trees for 30, 50, and 81 languages against the Ethnologue reference, $E$, for trees reconstructed by UPGMA and NJ for each combination of parameters described in Section~\ref{sec:methods}. Each dot represents a single reconstructed tree, $T$, and a tree distance, $D$. We performed 100,000 random permutations of the leaves of $T$, and compared each to the reference $E$ using the distance $D$. This yields a distribution with mean $\mu$ and standard deviation $\sigma$. To evaluate the reconstruction $T$, we plot $(\mu - D(T,E)) / \sigma$. The higher the value, the better the reconstruction. A star inside a dot signifies that $D(T,E)$ is smaller than 95,500 of the permuted tree distances.}
    \label{fig:sd-analysis-tda-tree-permutations}
\end{figure*}

\subsection{Evaluating Phylogenetic Trees} \label{sec:methods:evaluating_trees}

To compare the reference Ethnologue tree $E$ to each phylogenetic tree $T$ constructed using the language distances inferred by TDA, we employ six different tree distances, using implementations from the \texttt{R} package \texttt{TreeDist} version \texttt{2.6.3}.
These include
(i/ii) Jaccard-Robinson-Foulds distance for $k=1$ and $k=2$ \normalfont{\cite{Nye2005, BoCaKl2013}},
(iii) matching split distance \normalfont{\cite{BoGi12, LiRaMo12}},
(iv/v) the phylogenetic and clustering information distances \normalfont{\cite{Smith2020}}, and
(vi) path distance \normalfont{\cite{Fa69, StPe93}}.
We provide detailed descriptions of these distances in Appendix Section~\ref{sec:tree_distances}.

To assess whether the reconstructed trees capture part of the real phylogeny, we evaluate them in terms of leaf label permutations---comparing the distance $d(T,E)$ between the algorithmic tree $T$ and reference $E$ to the distribution of $d(T',E)$, where $T'$ is obtained from $T$ by shuffling the leaf labels.

For each of the 144 constructed trees, $T$, we performed 100,000 permutations of its leaf labels. For each permuted tree, we measure the six tree distances to the reference Ethnologue tree with the corresponding number of leaves---30, 50 or 81. Then, separately for each of the six tree distances, we identify where in this distribution of distances $d(T,E)$ lies. The measure of success is to check how many random permutations did worse than $T$.

\section{Results of the Experiments}
\label{sec:experiments}

The experiments show correlation between the word embedding dissimilarity and language phylogeny---for 484 out of 864 distributions, at least 95,000 (95\%) of the permutations are further from the reference tree $E$ than the reconstructed tree $T$ is.

Figure~\ref{fig:sd-analysis-tda-tree-permutations} is the primary result in this paper, presenting the results of evaluations of 144 different tree reconstructions based on TDA, each evaluated on six different tree similarity measures, for a total of 864 different points.
By employing permutation tests, we can map each such point to a significance level, expressed in terms of (a) the fraction of random samples dominated by the reconstructed tree or (b) the number of standard deviations the reconstructed tree sits from the mean value over 100,000 permutations of leaves; the options are tightly correlated. Figure~\ref{fig:sd-analysis-tda-tree-permutations} summarises~(b); analogous figure for (a) is in Appendix Figure~\ref{fig:sd-analysis-tda-tree-permutations_p-values}.

\begin{figure}[hbt]
    \centering
    \includegraphics[width=.45\textwidth]{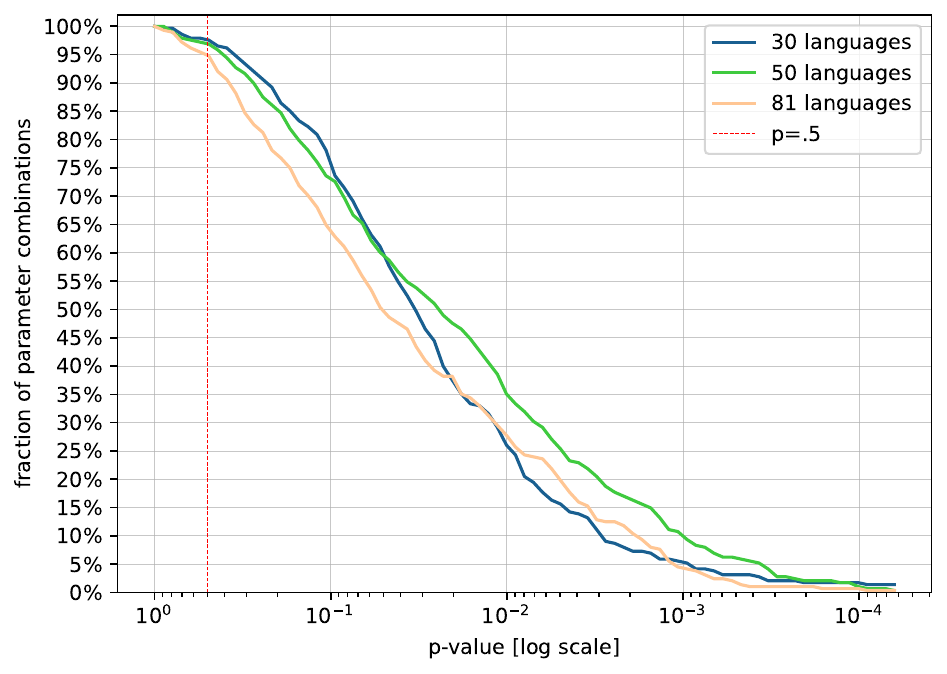}
    \vspace{-1ex}
    \caption{The fraction of parameter combinations (out of 288) in Figure~\ref{fig:sd-analysis-tda-tree-permutations} bested by $ \leq p\cdot 100,\!000$ permutations.}
    \label{fig:cumulative_p-val}
    \vspace{-2ex}
\end{figure}

Further summary of these results is in Figure~\ref{fig:cumulative_p-val}.
The cumulative distribution plots show the fraction of experimental conditions which satisfy a given permutation test $p$-value.
With respect to this value, 484 out of the 864 conditions proved significant to at least the $0.05$-level.
Of these, 255 were significant to at least $0.01$, 57 to at least $0.001$, and 10 to $0.0001$. In one case the algorithmic tree dominated all $t=100,\!000$ random permutations evaluated.

Our statistically strongest results came when using $50$ languages: possibly large enough to encode topologies hard to find by chance while avoiding low resource languages with poorly shaped embeddings.
Twice as many tests for UPGMA trees (114) were significant to the $0.005$-level as NJ trees (63), demonstrating that better combinatorial optimization leads to more significant phylogenies.

Different distances between persistent diagrams seem to fit well with different degree and token-to-token metric. Overall, bars statistics and sliced Wasserstein seem to perform better, while bottleneck is the least successful---this might be explained by its strong sensitivity to outliers, and suggests that the other notions might be better suited for analyses of word embeddings' (dis)similarities. Detailed breakdowns of the success of different parameters are provided in Tables~\ref{tab:aggregated_results_std}, \ref{tab:aggregated_results_geq99500}, \ref{tab:dots_greater_than_sigma}, \ref{tab:dots_in_top_percentage} in the Appendix.

That our best tree beats all trial $t$ permutations in our experiment implies it is significant beyond the $1/t$ level.
By measuring the quality score of each tree with respect to each distance in terms of the number of standard deviations above its random background mean and assuming normality, we can obtain a $p$-value associated with each such $z$-score.
The maximum achieved $6.87 \sigma$ corresponds to a Bonferroni-corrected $p$-value of $2.77 \times 10^{-9} = P\left[x\geq 6.87\right] * 864$, assuming a normal distribution. 
Coupled with all other presented permutation test results, clearly TDA is picking up real signal from the shape of word embeddings.

\section{Conclusions}\label{sec:conclusions}

Our analysis of FastText word embeddings of 81 Indo-European languages suggests that deviations from pair-wise isometry correlates with the historical origin of the languages---an interesting contribution to the debate about the extent to which the variance between monolingual embeddings is explained by essential differences between the languages versus scarcity of resources and under-training when constructing them.

A natural question is to what extent the presented results generalize. Would we get comparably significant results for word embeddings trained with different algorithms, with different parameters or on a different corpus? And if more careful training should lead to isometric embeddings, as suggested by \citet{vulic2020-good-isomorphic}, is there an explanation for a correlation of lower quality of embeddings with language phylogeny?  

The results suggest that TDA methods can help us better understand the structure of word embeddings, and further research can lead to useful insights. Different parameters can be better suited for comparing word embeddings than the previously used bottleneck lower bound of Gromov-Hausdorff distance. Furthermore, better understanding of the particular detected topological features could lead to interesting insights. For example, looking at a single language embedding, we can look back at concrete loops in degree~1 persistence diagrams and ask whether the collection of words it passes through has some meaningful linguistic interpretation. Or, identifying a large loop in one language, we can ask whether the translation of the words also forms a large loop in other languages.

\vfill
\pagebreak

\section*{Limitations}

In this paper, we have demonstrated that TDA methods are capable of reconstructing a real (albeit somewhat weak) signal about language structure and history from unlabeled word embeddings.

However, there are reasons to be optimistic that the strength of this signal may be increased with further research and experience.
We note that our TDA analysis is ``blind'', with no statistical normalizations of numerical range across the different language embeddings.
Orthogonal transformations preserve distances and angles, so they do not influence persistent homology, but scaling does change Euclidean distances, and translation (shifts) can change the cosine distances.

Outlier points may impact persistent homologies in misleading ways. For example the ``U-'' token in the English language embedding skewed the Euclidean 0-dim homology, even though it was completely isolated.
This motivates the question of whether we can preprocess data to eliminate clear outliers before analysis.

There is somewhat of a gap between the low-dimensional persistent homologies of dimensions $0 \leq d \leq 2$ which are computationally accessible and the 300-dimensional word vectors we analyze. In hopes of capturing more of the higher order behavior of high-dimensional data, data is often first reduced to fewer dimensions before persistent homology is applied. Standard methods like PCA, t-SNE or U-Map can be used to reduce dimensionality to $\sim 10$ or $\sim 50$.

We restrict our attention here to one popular but particular set of GloVe embeddings, without evaluating other methods such as word2vec, which are geometrically quite different and may in principle have even better topological properties.
There are also reasonable questions of whether our results may strengthen if we used a different reference tree than provided by Ethnologue which is difficult to compare to: it is unrooted, non-binary, and very flat.

The phylogeny of languages might also not be the aspect most correlated with the distances we see. For example geographical and structural similarities as studied in \citet{bjerva2019-representations} could lead to closer correlations.

Our results demonstrate that TDA invariants capture properties of languages, but what aspects of language are they keying on?
What do tunnels and voids in embeddings tell us about languages?

\newpage

\bibliography{anthology,bibliography}{}

\newpage
\ 
\newpage

\appendix

\section{Supplemental Material}

\subsection{Vietoris-Rips complex}\label{sec:vietoris-rips}

To compute persistent homology, the continuous spaces of growing balls around the points in the cloud need to be replaced by a discrete structure easily represented in a computer. Concretely, by a simplicial complex, a (hyper)graph-like structure. A \emph{simplicial complex} consists of a set of vertices, and a set of faces. Each face is a subset of vertices, and for each face, a simplicial complex has to also include all its subsets. A \emph{filtration} of a simplicial complex is a function that assigns a real number to each face, and is monotone with respect to inclusion, that is, a subset of a face has to have smaller or equal value. A collection of all faces with value at most $r\geq0$ is then a simplicial complex. This yields a chain of growing simplicial complexes.

For low-dimensional Euclidean data, persistent homology is usually computed using Delaunay complexes with alpha filtration. From the topological perspective, those complexes exactly reflect the spaces of growing disks. However, computing Delaunay complexes is not practically viable for data in dimensions beyond $\sim 10$. A widely used alternative is the Vietoris-Rips complex. It can be constructed from any distance matrix, and for Euclidean data offers a good approximation of the alpha filtration. Its low-dimensional persistent homology can be computed efficiently, and a fast implementation is readily available.

Given an $n\times n$ distance matrix $D$, and a radius $r>0$, the \emph{Vietoris-rips complex}, $\VR{r}{D}$, is the flag complex of all the edges $\{i,j\}$ with $D_{i,j}\leq r$. That is, $\sigma\subseteq [n]=\{1,\dots, n\}$ is a face of $\VR{r}{D}$ iff $D_{i,j}\leq r$ for all $\{i,j\}\subseteq\sigma$. As the radius grows to infinity, the complex grows to become the full-simplex $2^{[n]}$ whose size grows exponentially with respect to $n$. The reason that this is not a problem for computations is that the full complex need not be computed to obtain persistent homology. Firstly, only edges, triangles and tetrahedra are relevant when we only care for $0$-, $1$- and $2$-dimensional persistent homology, secondly, there exists a radius cut-off after which the homology is guaranteed to be trivial, and finally, the relevant structures can be used implicitly rather than stored explicitly in memory. For more details see \cite{Bauer2021Ripser}.

For completeness we provide the formal definition of a `feature' within a VR complex.
A boundary $\partial\sigma$ of a face $\sigma\in\VR{r}{D}$ is the collection of subsets $\sigma\setminus\{i\}$ for each $i\in\sigma$. An addition operation is defined on boundaries as the symmetric difference of the collections. A boundary of a collection of faces is then the sum of their boundaries. For example, the boundary of two triangles sharing an edge are the four edges on the outside.

There is a bar $[r_b,r_d)$ in the $p$-dimensional persistence diagram of $\VR{\bullet}{D}$ iff there exist two collections of faces, $\mathcal{B}\subseteq\VR{r_b}{D}$ and $\mathcal{D}\subseteq\VR{r_d}{D}$, with the following properties. First, the boundary $\partial\mathcal{B}$ is empty, and one of the faces in $\mathcal{B}$ appears for the first time in $\VR{r_b}{D}$. Second, the boundary $\partial\mathcal{D}$ is $\mathcal{B}$, and one of the faces in $\mathcal{D}$ appears for the first time in $\VR{r_d}{D}$. For example, a $1$-dimensional bar means that an edge appears in $r_b$ which completes a loop, and then this loop is filled by a collection of triangles, the last of which appears in $r_d$.

\subsection{Preserving the Ethnologue Tree Topology}
\label{sec:preserving-ethnologue-topology}

Our previous experiments compared algorithmically-constructed trees (UPGMA and NJ) built using TDA-based distance matrices against the Ethnologue reference trees on identical sets of languages.
A possible objection to this approach is that the reference trees are generally much shallower than those constructed by the algorithms, perhaps affecting our evaluation.

Thus we investigated an optimization procedure to construct TDA-based trees consistent with the reference tree topology.
 Instead of constructing a new tree, we try to find the best assignment of labels (languages) to the leaves based on the given distance matrix. A tree always yields a distance matrix between the leaves---the distance of two leaves is the length of the unique path connecting them. We set up an optimization process to find the label assignment leading to the tree distance matrix that best correlates with our given distance matrix. To compare distance matrices, we use the Pearson correlation coefficient. Therefore, if $E$ is the tree distance matrix of the Ethnologue tree, and $D$ is the distance matrix computed from the pipeline as described above, we want to find a permutation matrix $P$ that maximizes the Pearson correlation coefficient between $P^TEP$ and $D$.

This is, however, a difficult optimization problem. To compute the Pearson correlation, the entries in the matrices are first replaced by their ranks, leading to matrices $E', D'$, and the coefficient is then the covariance $\mathrm{cov}(E', D')$ divided by the standard deviations. The collection of entries does not change when we permute the matrices, so we retain the same ranks, means, and standard deviations. This means that we search for $P$ maximizing $\langle P^TE''P,\, D'' \rangle_F$, where $E'', D''$ are the matrices whose entries are ranks minus the mean of ranks, and $\langle \cdot, \cdot \rangle_F$ is the Frobenius inner product, which is the standard scalar product of the matrices viewed as unraveled vectors. This is, however, an instance of the quadratic assignment problem, which is generally NP-hard.

Our search for an optimal permutation is a naive flip-heuristic. Starting with a random permutation, we try random transpositions, and keep them if they improve the correlation. This procedure gets stuck in a local optimum after a few thousand flips. For each distance matrix, we ran this process 100 times and took the maximum of the local optima.

\begin{figure*}[ht]
    \centering
    \includegraphics[width=1\linewidth]{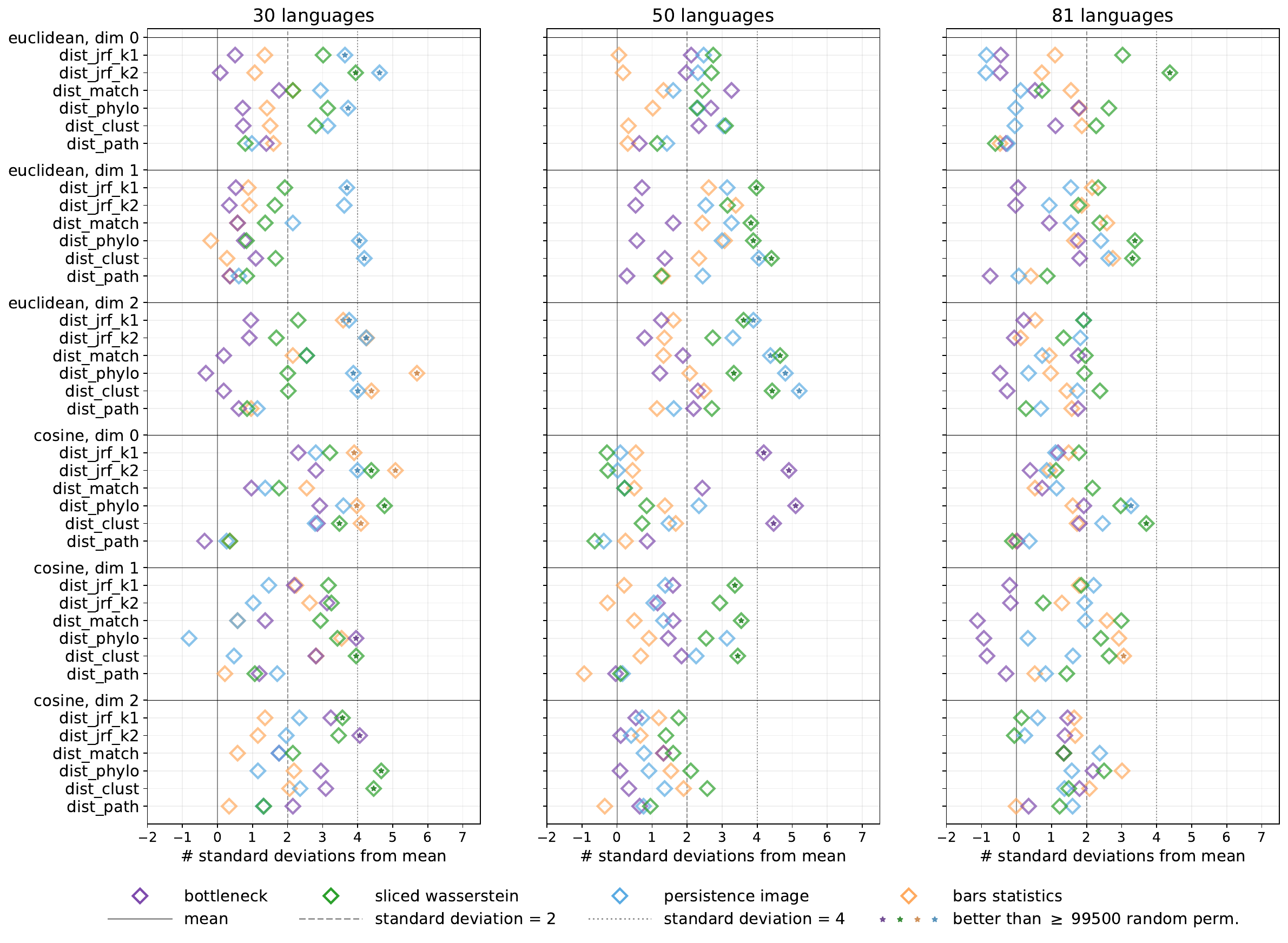}
    \caption{The statistical significance of the labelings of the Ethnologue tree that optimize distance matrix correlation (see Section~\ref{sec:preserving-ethnologue-topology}), for each combination of parameters described in Section~\ref{sec:methods}. Each dot represents the distance of a single reconstructed labeling to the reference Ethnologue tree, and its position shows how many standard deviations away from the mean it lies in a distribution of 100,000 random labelings of the Ethnologue tree.}
    \label{fig:sd-analysis-distance-matrix-correlation}
\end{figure*}

Our results are presented in Figure \ref{fig:sd-analysis-distance-matrix-correlation}.
Consistent with the results reported previously, we perform substantially better than chance for almost all combinations of TDA variant and tree distance metric, typically by two or more standard deviations from the mean.

\subsection{Tree Similarity Metrics}\label{sec:tree_distances}

To evaluate the experimental results, we need to quantitatively compare phylogenetic trees built using topological data analysis against the reference language tree from Ethnologue.
Many labeled tree-to-labeled tree distances have been defined in the literature. We chose six different distances in our analysis, considering meaningfulness in our setting, availability of implementation, and efficiency of computation.

{
Generally there are two main directions in comparing trees. One is to define some notion of distances between pairs of leaves---e.g., the length of the shortest path connecting leaves or the distances of the closest common ancestor to either the root or to the leaves---and compare the distance matrices. This approach is not particularly well suited for our application, as reference Ethnologue tree is very flat. Nevertheless, we include one such distance.
}

\paragraph{Path distance \normalfont{\cite{Fa69, StPe93}}.} A path distance matrix of a tree is labeled by the leaves, and an entry is the length of the unique path in the tree that connects the corresponding two leaves. The distance of two trees is then the Frobenius distance of their path distance matrices---that is, the Euclidean distance when the matrices are unraveled into vectors.

\begin{figure*}[ht]
    \centering
    \includegraphics[width=.25\textwidth, valign=t]{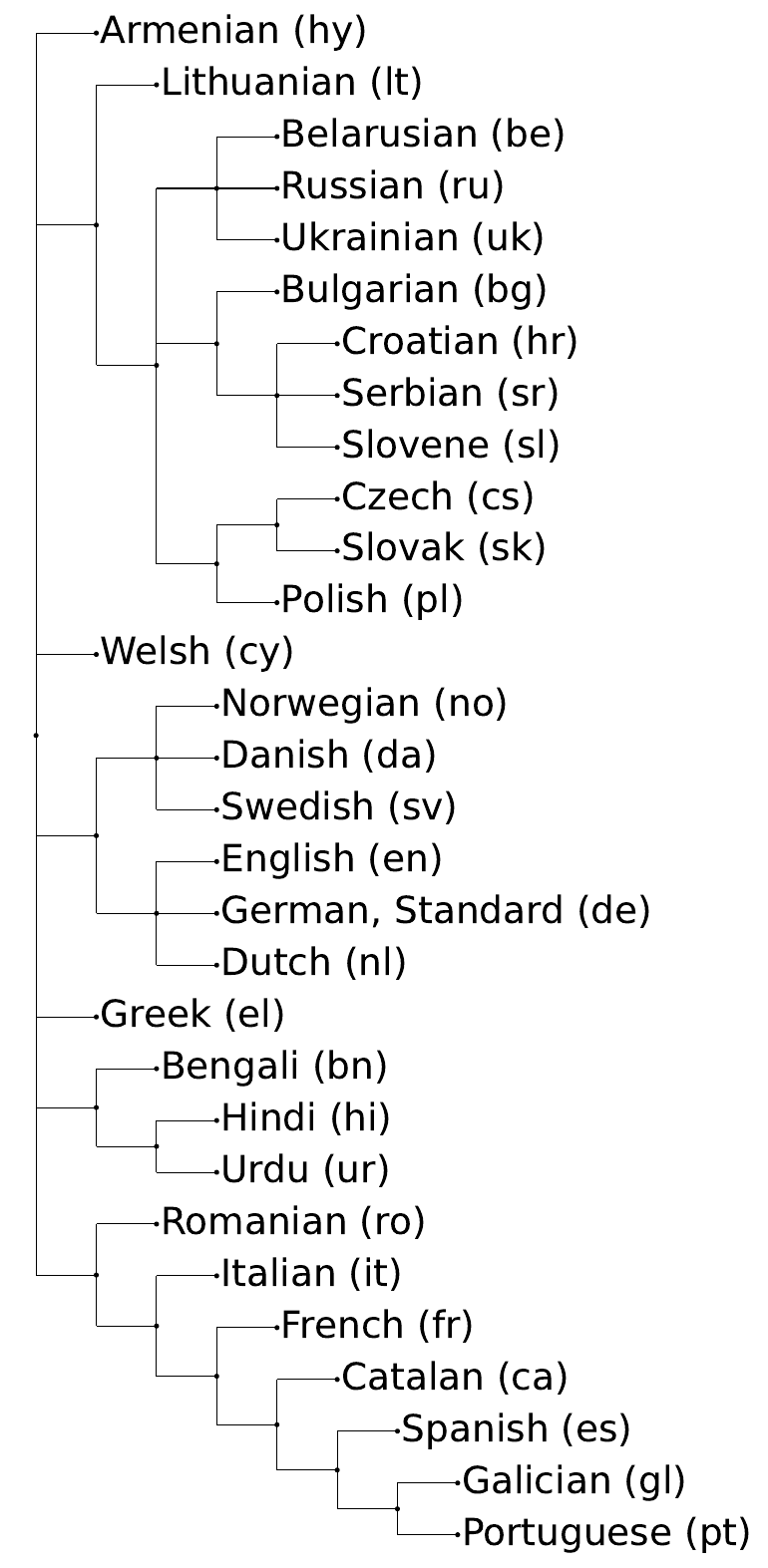}
    \hfill
    \includegraphics[width=.25\textwidth, valign=t]{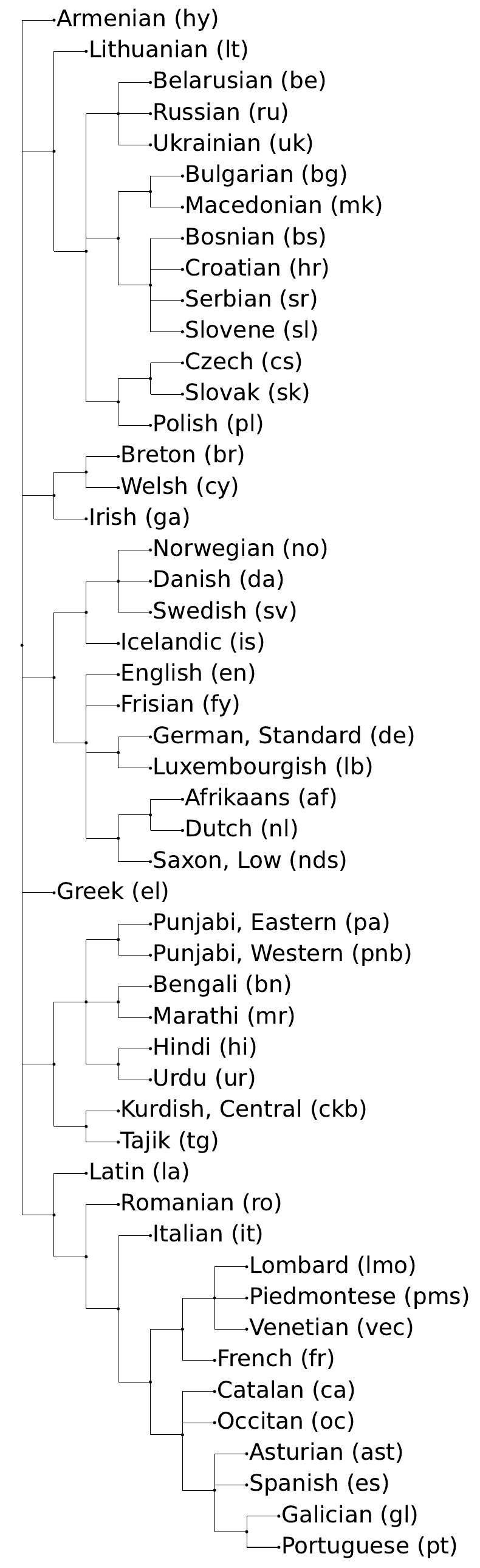}
    \hfill
    \includegraphics[width=.25\textwidth, valign=t]{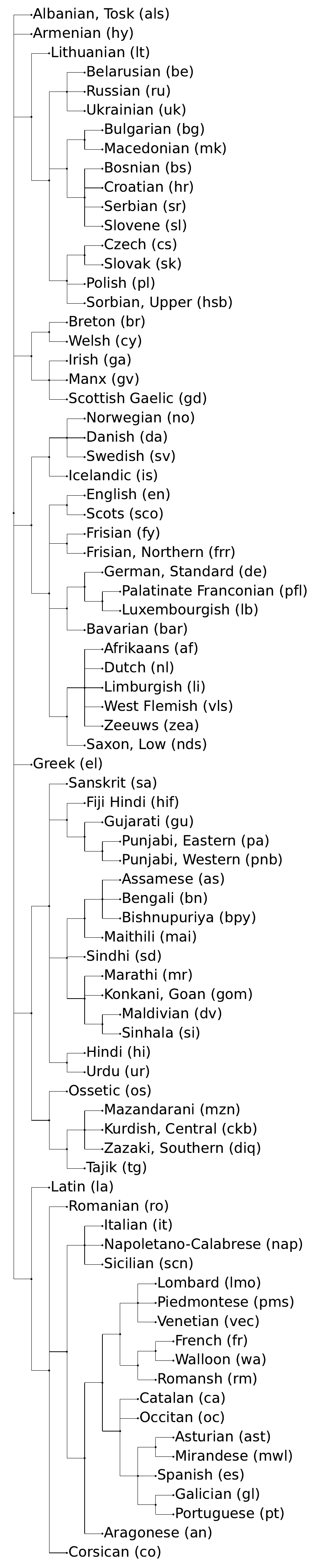}
    \caption{The reference Ethnologue language phylogenic trees \cite{ref:ethnologue} employed in this study, for sets of the most popular 30, 50 and 81 Indo-European languages with FastText word embeddings.}
    \label{fig:ethnologue_tree}
\end{figure*}

{
\medskip
The other option is to compare the different ways that the tree can partition its leaves into two groups by removing an edge. The first and simplest such method is the Robinson-Foulds (RF) metric \cite{robinson-foulds}. The distance of two trees, $T_1$, $T_2$, with the same leaves, is the number of partitions achievable by $T_1$, but not $T_2$, plus the number of partitions achievable by $T_2$ but not $T_1$; possibly normalised to range from $0$ to $1$. Partitioning is intuitively a good way to measure success of our method---after all, when naively comparing a constructed tree to the reference, we argue with claims like ``Slavic languages are well separated from the others''. The problem with RF is that comparing partitions on the nose is too penalizing. For example with a single swap of Portuguese with Czech in the 30-language Ethnologue tree (Figure~\ref{fig:ethnologue_tree}), we go from distance zero to more than half of the maximum. Even reasonably similar trees can often get the maximum distance.

To fix the issue, generalizations of RF metric quantify the similarity of any pair of partitions. To compute how similar two trees are, we find the matching of the partitions defined by $T_1$ with the ones defined by~$T_2$ that maximizes the sum of similarities of the matched partitions. To go from similarity to distance, we take the difference of theoretically maximal similarity and the computed similarity, and normalize to fit in the range $0$ to $1$. All the distances below follow the same logic, but differ in the definition of the similarity between two partitions. For the explanations, consider $A_0, A_1$ a partition of the leaves obtained by removing an edge from the tree $T_1$, and $B_0, B_1$ another partition arising the same way from $T_2$.
}

\paragraph{Jaccard-Robinson-Foulds distance \normalfont{\cite{Nye2005, BoCaKl2013}}.} JRF is a family of distances with one parameter, $k$. Its name comes from using the Jaccard index: $J(A_i, B_j)\coloneqq \card{A_i\cup B_j} / \card{A_i\cap B_j}$. The similarity of two partitions is defined as the bigger of the two values, $\min\left( J(A_0, B_0), J(A_1, B_1) \right)$ and $\min\left( J(A_0, B_1), J(A_0, B_1) \right)$, raised to the power $k$. As $k$ tends to infinity, JRF distance converges to RF. In our analysis, we use $k=1$ and $k=2$.

\paragraph{Matching split distance \normalfont{\cite{BoGi12, LiRaMo12}}.} The matching similarity of two partitions is the maximum of $\card{A_0\cap B_0}+\card{A_1\cap B_1}$ and $\card{A_0\cap B_1}+\card{A_1\cap B_0}$. The idea is that the number of leaves minus the similarity is the number of leaves that need to be swapped in order to obtain identical partitions.

\paragraph{Phylogenetic and Clustering information distances \normalfont{\cite{Smith2020}}.} These metrics use notions from information theory to quantify similarity of partitions. The former measures the information in bits shared between the two partitions, while the latter quantifies the similarity with shared entropy.

{
\medskip
For further details on both the intuition and the technicalities of the tree distances, see the documentation\footnote{\url{https://ms609.github.io/TreeDist/reference/index.html}} of \texttt{TreeDist} package for \texttt{R}, which we used for the computations.
}

\subsection{P-values from Permuting Labels} \label{sec:experiments:permuting_labels}

Although there exist many distance metrics for comparing rooted, labeled trees, these metrics can be difficult to interpret because they conflate differences in tree topologies (e.g height and width) with differences in leaf labelings (e.g. which label pairs are siblings of each other).
In our experiments with language phylogenies, we observed that the toplogies of trees built from different heuristics (NJ and UPGMA) differed dramatically, and that both differed substantially in shape from both the reference tree and random trees constructed from natural agglomerative processes.

To evaluate the statistical significance of TDA-inspired trees, we propose the following procedure to mitigate impact of differing topology on our distance metrics.
To evaluate how well an algorithmic tree ($T$) agrees with a given reference tree ($R$) with respect to a given distance metric $d(T,R)$, we construct a set of $n$ random label-permuted algorithmic trees $T'(i)$, $1 \leq i \leq n$.
These label-permuted trees will each have the exact same topology as $T$, but the leaf-labels of $T$ be randomly and independently permuted in each $T'(i)$.

Comparing $d(T,R)$ to $d(T'(i),R)$ reflects only the differences in labeling while conserving topology. The degree to which $T$ is better than the background of label-permuted can be assessed in two different ways.
First, the rank $r$ of the value $d(T,R)$ against the universe of $n$ label-permuted trees yields a p-value of $(r-1)/n$.
Second, after computing the mean $\mu$ and standard deviation $\sigma$ of the distribution of $d(T'(i),R)$, we can interpret $d(T,R)$ in terms of the number of standard deviations from the mean, which map directly to p-values assuming a normal distribution. We employ both techniques in the analysis to follow. The Spearman correlation of the two approaches is larger than $0.99$, which indicates that we can use them interchangeably. We prefer the latter as it also takes the spread of the distribution into account, and can additionally identify meaningful $p$-values of statistical significance greater than $1/100,\!000$.

\subsection{Significance Results}

\begin{figure}[ht]
    \centering
    \includegraphics[width=.45\textwidth]{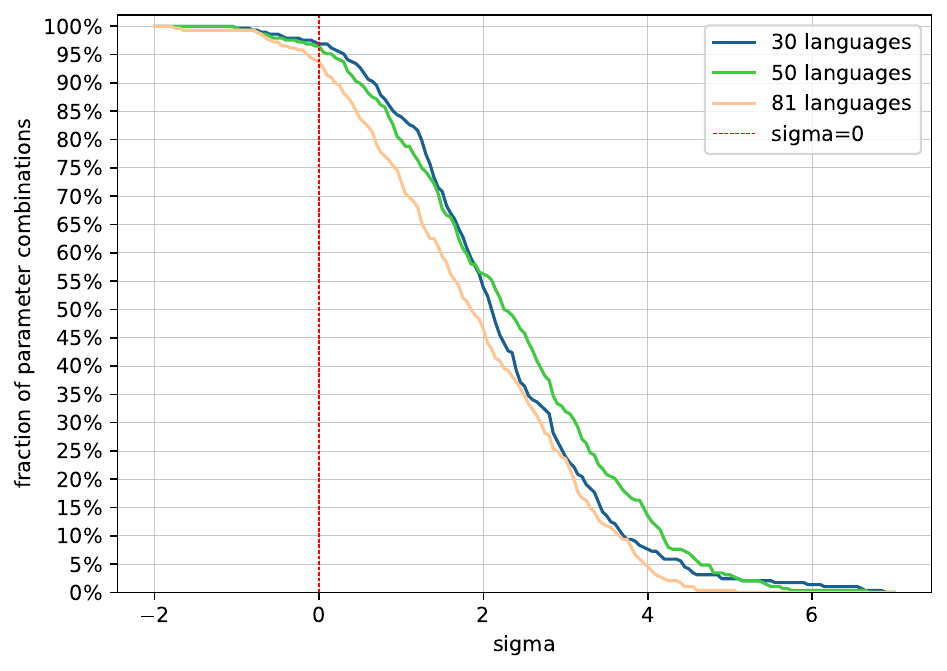}
    \caption{The fraction of parameter combinations (out of 288) in Figure~\ref{fig:sd-analysis-tda-tree-permutations} further away from the mean than $\sigma$ standard deviations.}
    \label{fig:cumulative_sigma}
\end{figure}

The result for the number of standard deviations from the mean are presented in Figure \ref{fig:cumulative_sigma}.
Fully 451 out of 864 conditions yielded results that sat at least $2 \sigma$ above the mean, with 229 at least $3 \sigma$, 74 at least $4 \sigma$, and five conditions at least $6 \sigma$ above the mean, peaking at $6.87 \sigma$. For detailed breakdown see Table~\ref{tab:dots_greater_than_sigma}.

\subsection{Evaluation of TDA Methods}

To better understand which parameters perform the best, we aggregate the information from Figure~\ref{fig:sd-analysis-tda-tree-permutations} in two ways. In Table~\ref{tab:aggregated_results_std}, we give the averages of standard deviations away from the mean when we fix some parameters and vary others, while in Table~\ref{tab:aggregated_results_geq99500} we show the numbers of dots that performed better than $99.5\%$ of the random permutations.
In the first block of each table we compare metric for the word embedding and persistent homology dimension to compare the different distances.
In the second block we compare the persistence diagram distance to compare how meaningful different topological features were.
In the third block we compare tree construction algorithms.

\begin{figure*}[th]
    \centering
    \begin{subfigure}{.24\textwidth}
        \includegraphics[width=\linewidth]{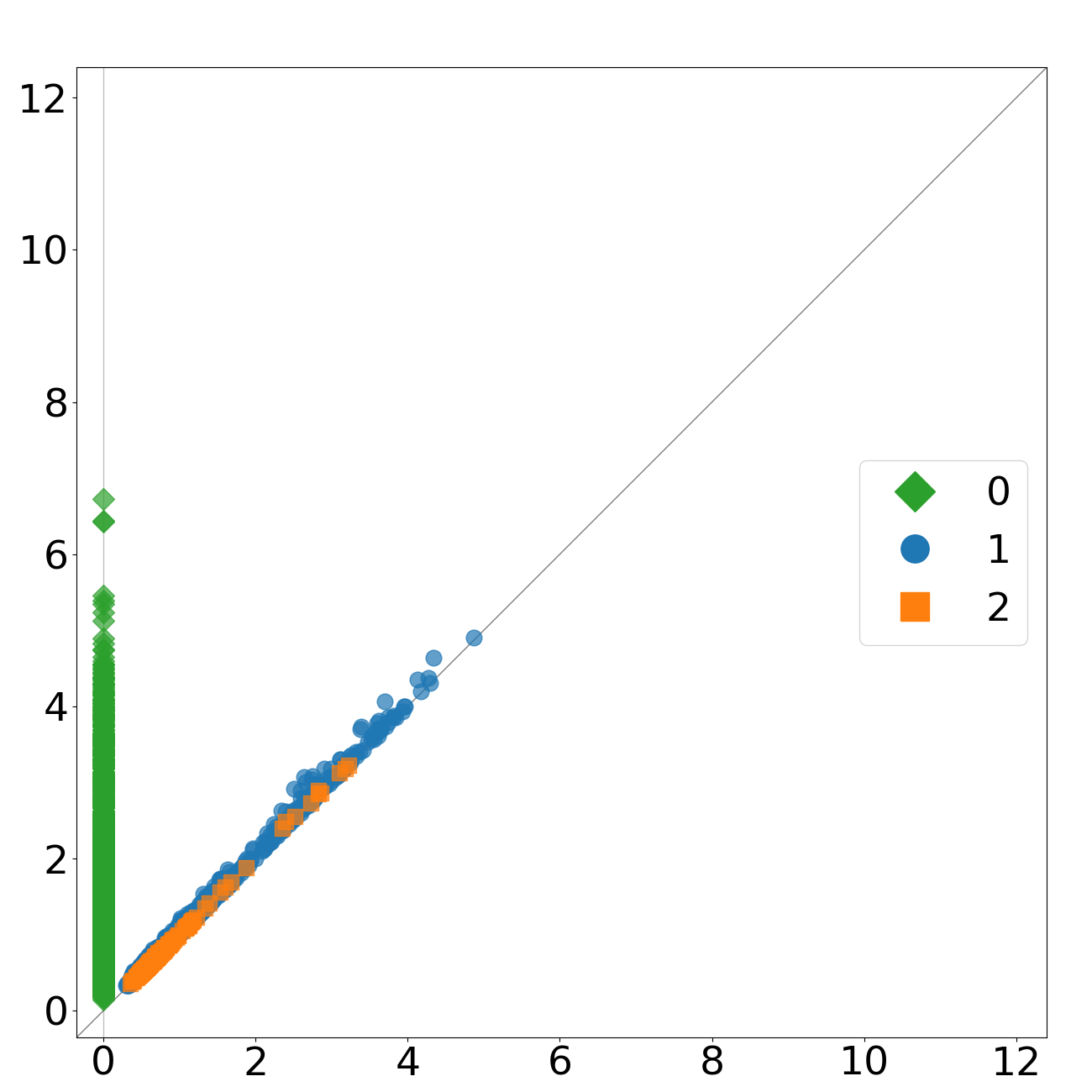}
        \caption{Czech (cs), Euclidean}
    \end{subfigure}\hfill
    \begin{subfigure}{.24\textwidth}
        \includegraphics[width=\linewidth]{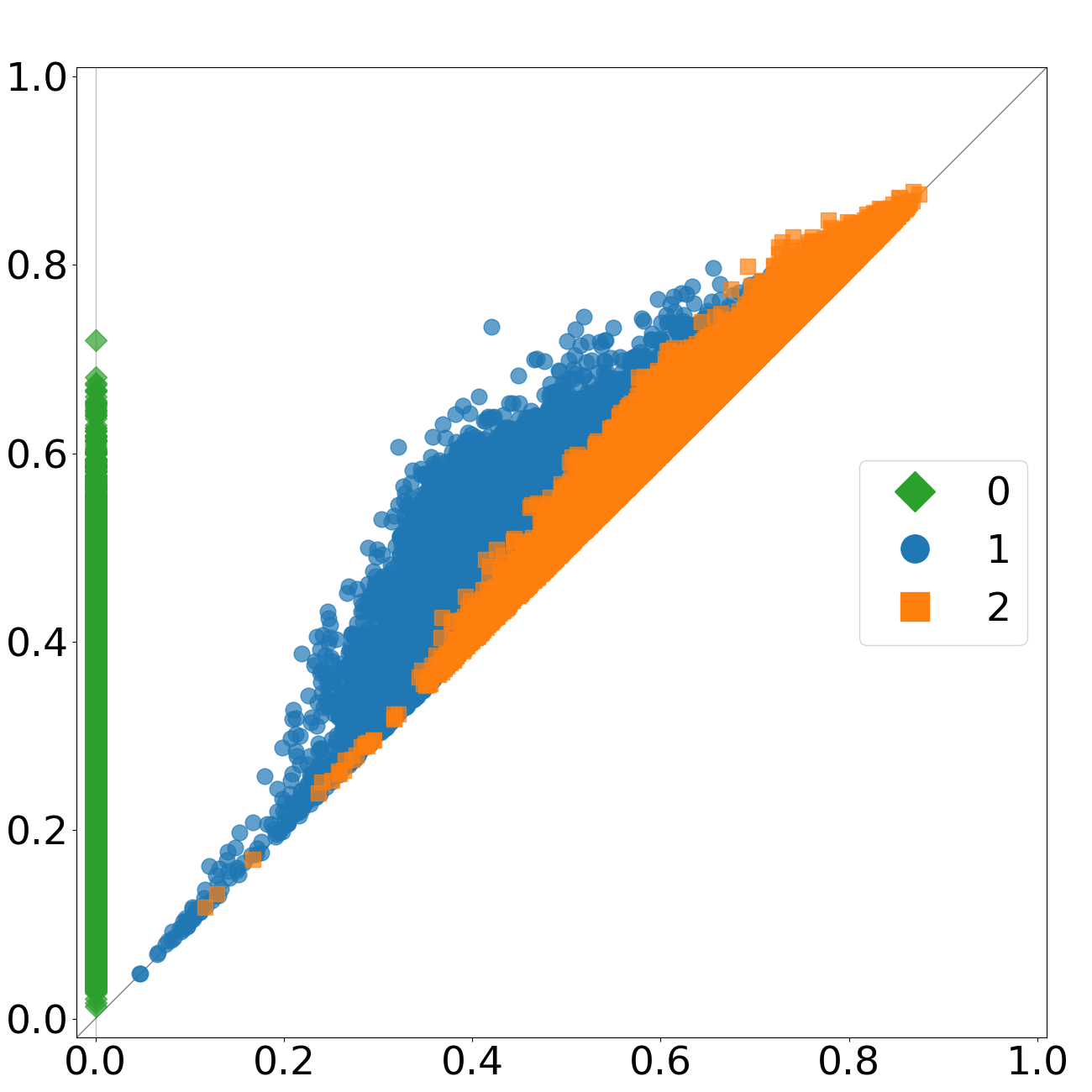}
        \caption{Czech (cs), cosine}
    \end{subfigure}\hfill
    \begin{subfigure}{.24\textwidth}
        \includegraphics[width=\linewidth]{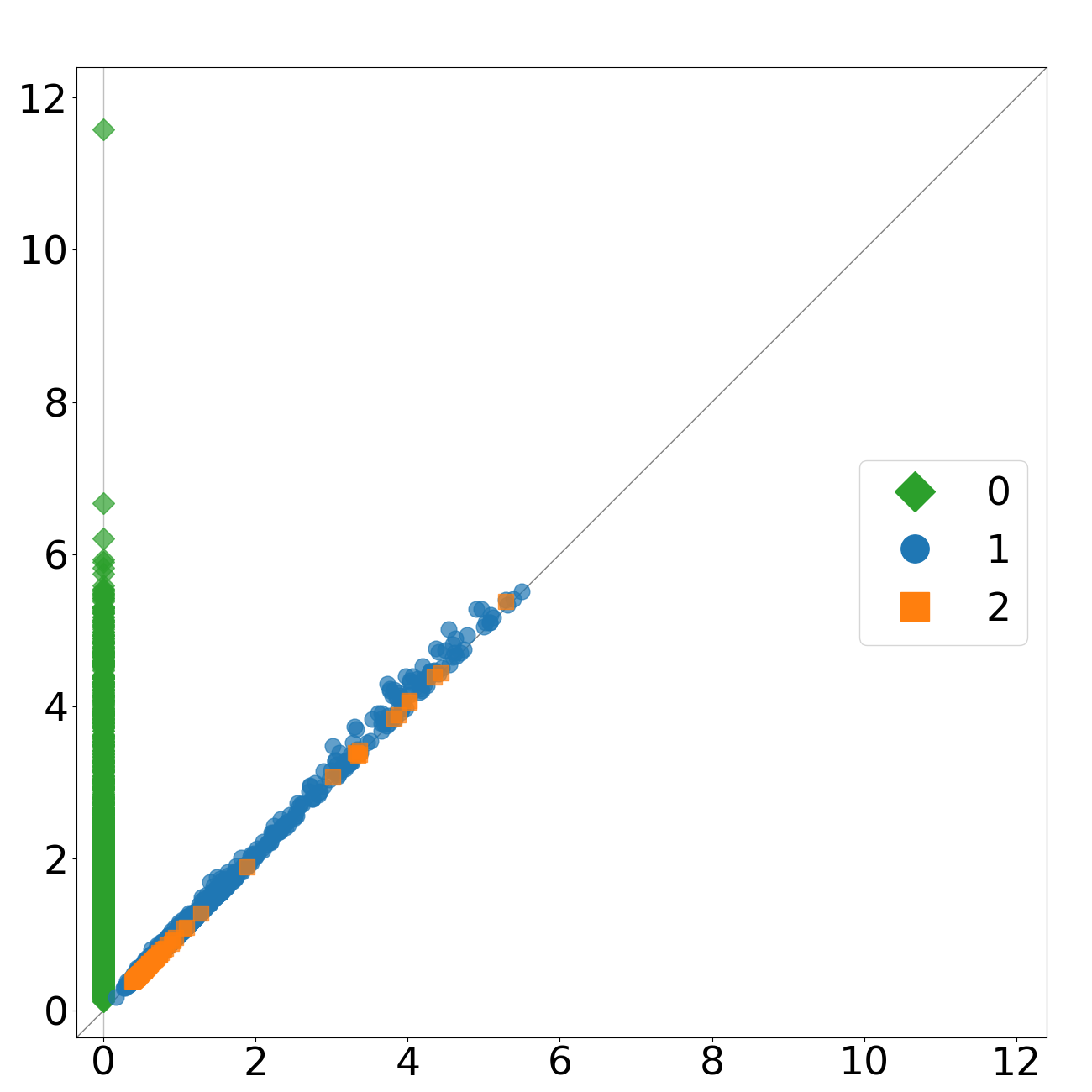}
        \caption{English (en), Euclidean} \label{fig:persistence_diagrams:en_eucl}
    \end{subfigure}\hfill
    \begin{subfigure}{.24\textwidth}
        \includegraphics[width=\linewidth]{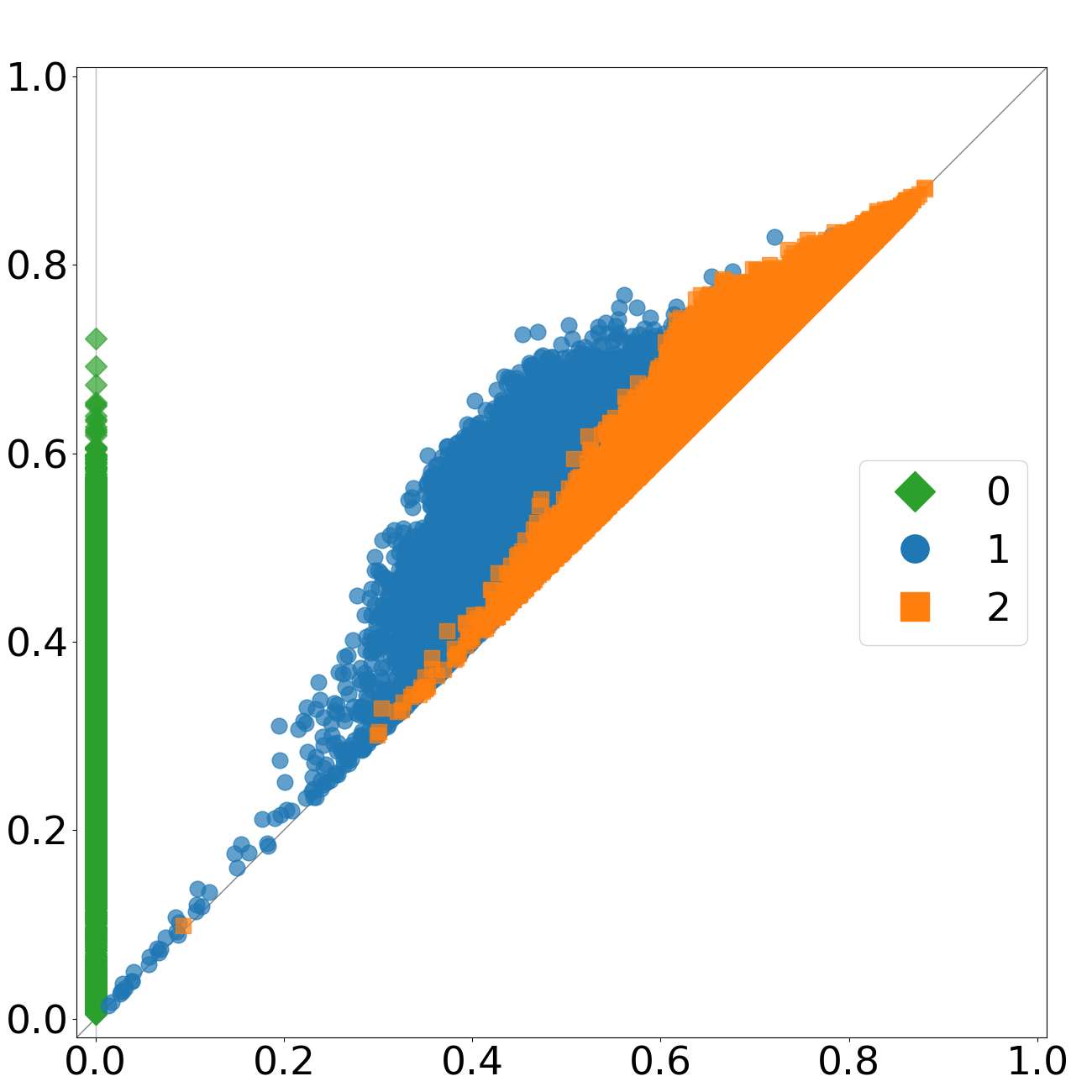}
        \caption{English (en), cosine}
    \end{subfigure}

    \smallskip
    \begin{subfigure}{.24\textwidth}
        \includegraphics[width=\linewidth]{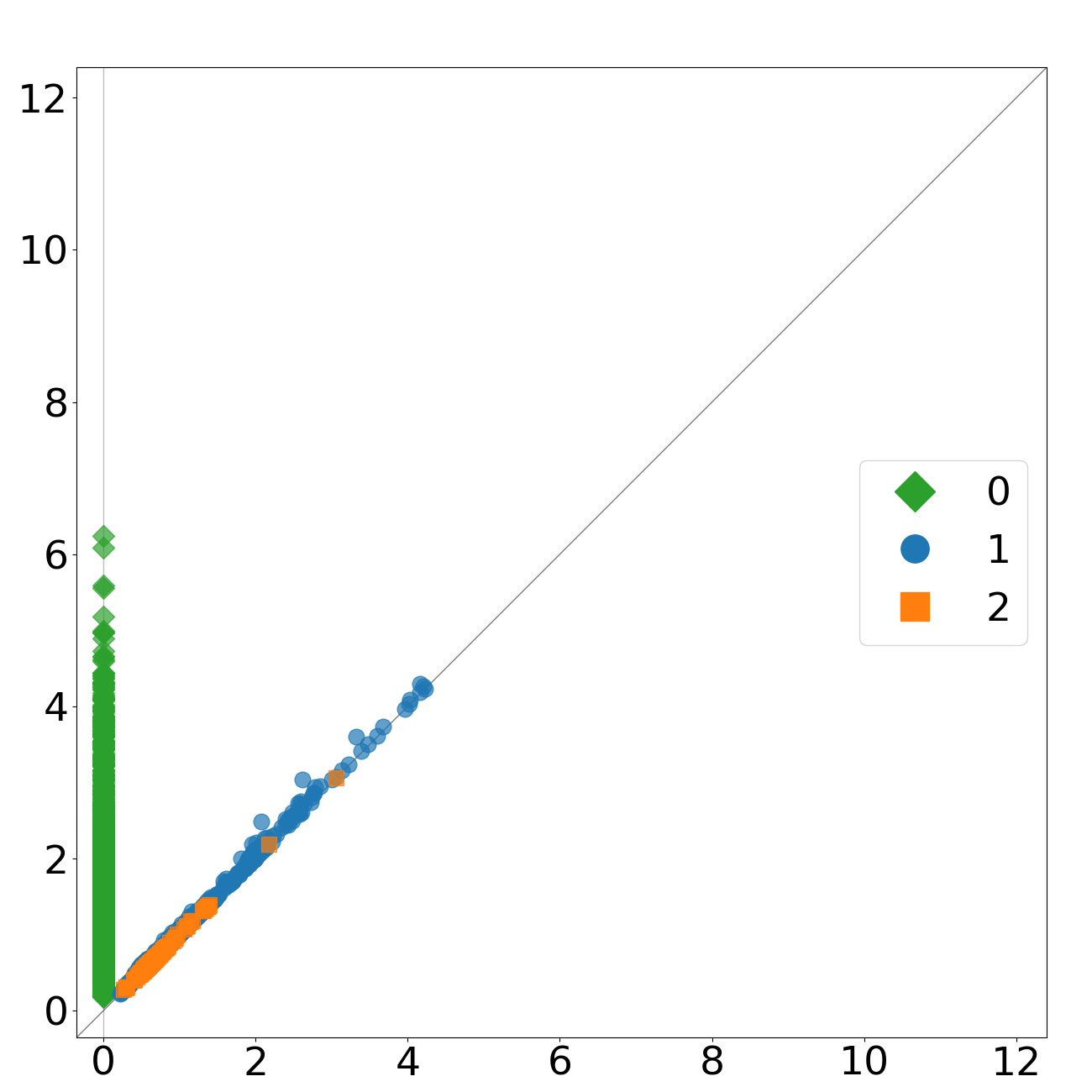}
        \caption{Slovak (sk), Euclidean}
    \end{subfigure}\hfill
    \begin{subfigure}{.24\textwidth}
        \includegraphics[width=\linewidth]{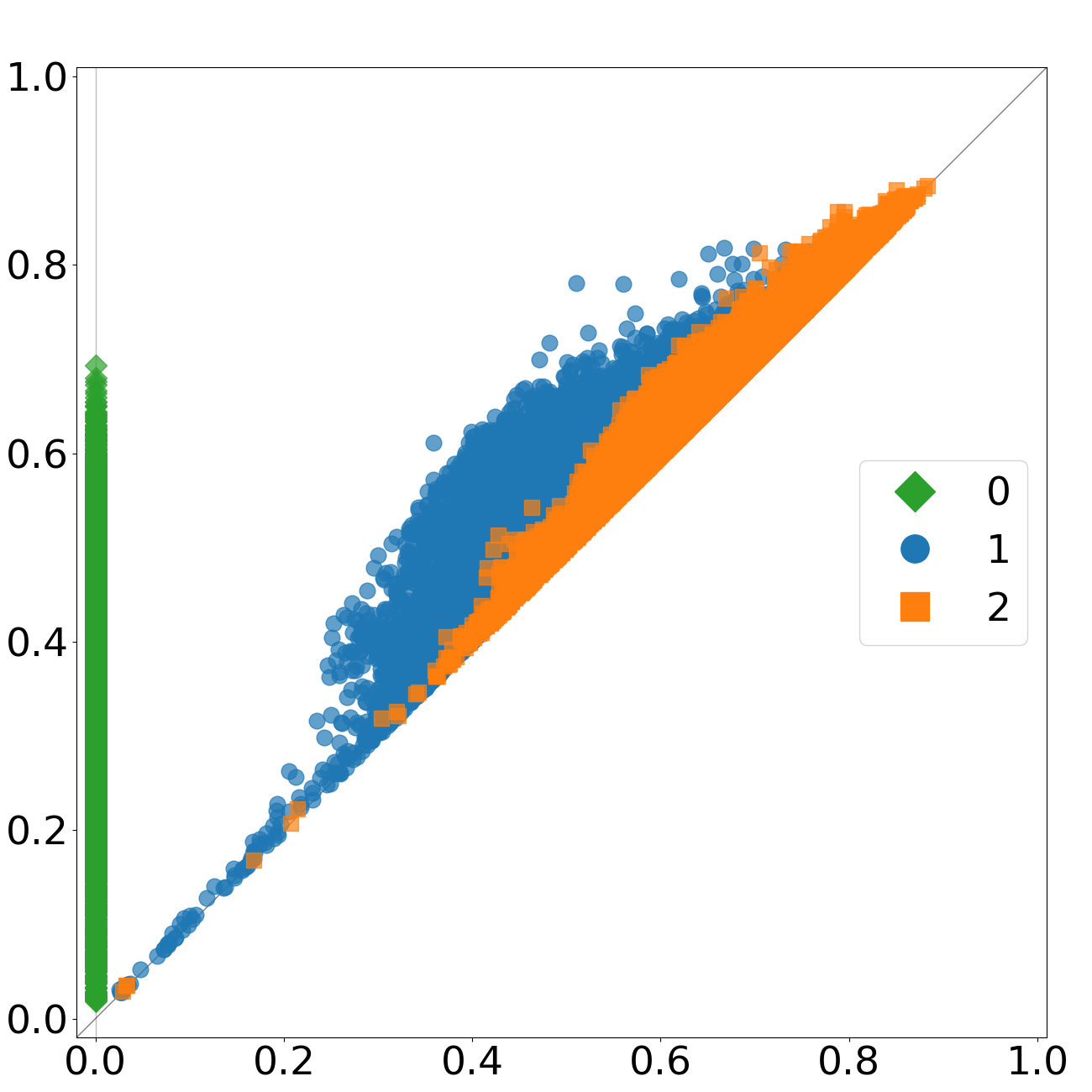}
        \caption{Slovak (sk), cosine}
    \end{subfigure}\hfill
    \begin{subfigure}{.24\textwidth}
        \includegraphics[width=\linewidth]{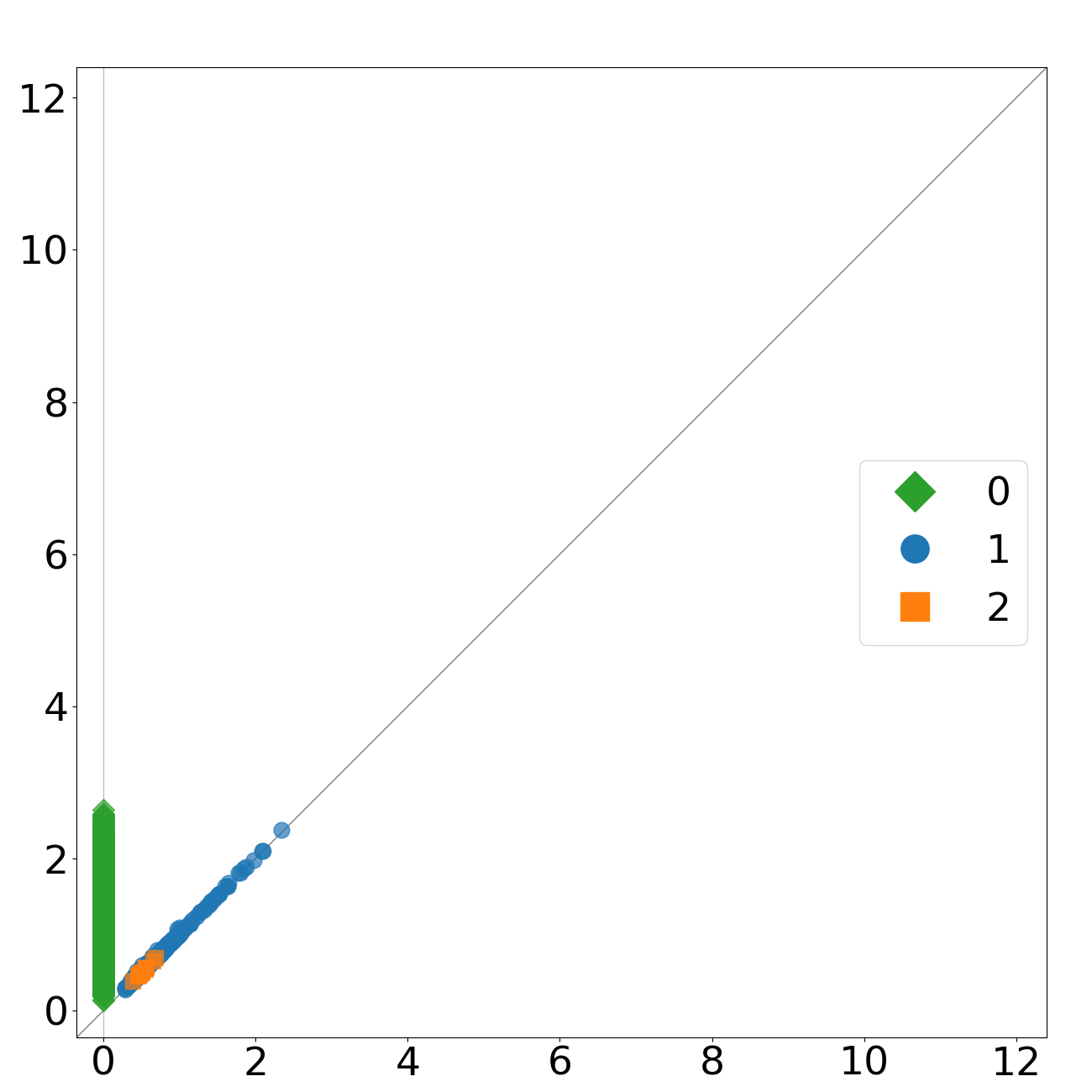}
        \caption{Irish (ga), Euclidean}
    \end{subfigure}\hfill
    \begin{subfigure}{.24\textwidth}
        \includegraphics[width=\linewidth]{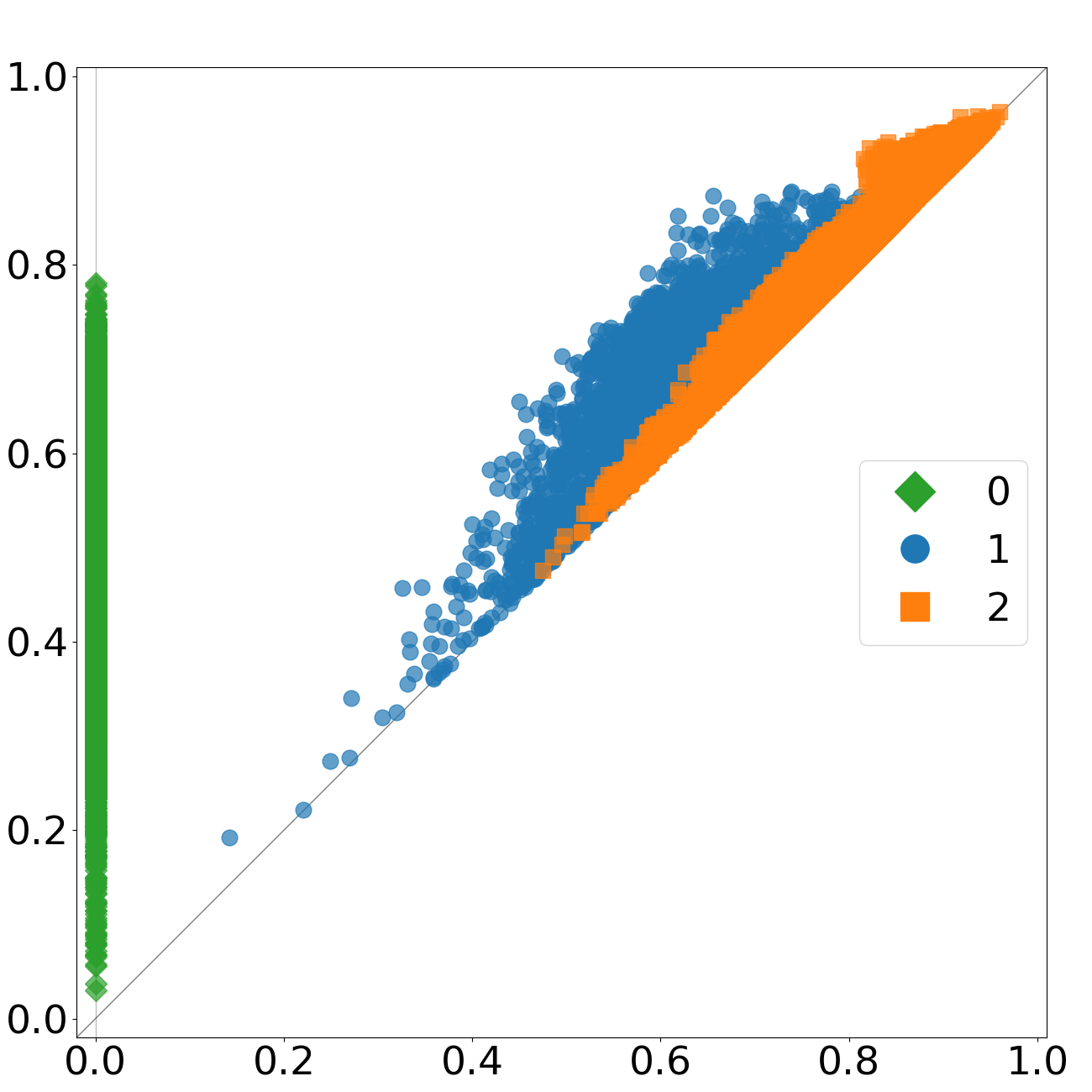}
        \caption{Irish (ga), cosine}
    \end{subfigure}

    \caption{Eight persistence diagrams for language embeddings analysed in this paper including Czech and Slovak, which are very closely related, and English and Irish, which belong to distant branches (Germanic and Celtic) of the Indo-European family. For each language we show diagrams for distinct Euclidean and cosine metrics.
    Each diagram shows features for dimensions 0, 1 and 2, that is, connected components, loops, and 2-spheres.}
    \label{fig:persistence_diagrams}
\end{figure*}

Overall dimension~1 seems to outperform 0 and 2 for both Euclidean and cosine metric, and there is no clear winner between the two metrics. It is, however, surprising that Euclidean metric with 2-dimensional topological features is not too far behind. When we look at the persistence diagrams on Figure~\ref{fig:persistence_diagrams}, we see that the 2-dimensional diagrams for Euclidean metric are somewhat sparse and all the features seem to be close to the diagonal. Indeed, probing all the diagrams, the maximum persistence of 2-dimensional features is roughly comparable to the smallest distance between the embeddings of any two words for the given language. Since features can pop up or disappear at the diagonal with small perturbations of data, such low-persistent features would often be discarded as noise. Yet, in combination with bars statistics distance, the trees constructed with UPGMA algorithm based on this data are among the best performing in our analysis.

From the persistence diagram distances, sliced Wasserstein and bars statistics lead to clearly stronger results than bottleneck distance and persistence images. Bottleneck distance is expected to be subpar, as having one more or one fewer feature can already yield a large distance.
It is a bit less clear why persistent images do not perform so well, especially when they should approximate $1$-Wasserstein distance just as the sliced Wasserstein. One possible reason is that the method has many parameters and our choice might not be ideal---we chose range, for both birth and death, $(0,1)$, $\sigma=.1$, for cosine; and $(0,10)$, $\sigma=1$, for Euclidean; grid $10\times 10$.

\newpage

\begin{table}[htb]
    \centering
    \scriptsize
    {
    \setlength{\tabcolsep}{4pt}
    \begin{tabular}{l|llllll|l}
        \# $\sigma$ from mean &  j\_k1       & j\_k2       & mat         & phy         & clust         & path          & mean           \\
        \hline &&&&&&& \\[-2ex]
        Euclidean, dim 0   &  2.19          & 2.54          & 1.37          & 2.14          & 2.19          & 0.83          & 1.88          \\
        Euclidean, dim 1   &  \textbf{2.74} & \textbf{2.57} & 2.33          & 2.74          & 3.10          & 1.14          & 2.43          \\
        Euclidean, dim 2   &  2.38          & 2.17          & 2.17          & 2.38          & 2.74          & 0.76          & 2.10          \\
        cosine, dim 0      &  2.19          & 2.19          & 1.53          & 2.63          & 2.58          & 0.62          & 1.95          \\
        cosine, dim 1      &  2.41          & 2.14          & \textbf{2.58} & \textbf{3.28} & \textbf{3.27} & \textbf{1.46} & \textbf{2.52} \\
        cosine, dim 2      &  1.72          & 1.36          & 1.76          & 2.89          & 2.71          & 0.93          & 1.90          \\[.8ex]
        \hline &&&&&&& \\[-2ex]
        bottleneck         &  1.80          & 1.70          & 1.57          & 2.28          & 2.36          & 0.93          & 1.78          \\
        sliced Wasserstein &  \textbf{2.63} & \textbf{2.49} & \textbf{2.40} & 2.95          & 3.12          & \textbf{1.05} & \textbf{2.44} \\
        persistence image  &  2.25          & 2.17          & 1.69          & 2.21          & 2.38          & 0.88          & 1.93          \\
        bars statistics    &  2.40          & 2.28          & 2.17          & \textbf{3.26} & \textbf{3.19} & 0.97          & 2.38          \\[.8ex]
        \hline &&&&&&& \\[-2ex]
        nj                 &  2.20          & 2.14          & 1.70          & 2.48          & 2.65          & 0.90          & 2.01          \\
        upgma              &  \textbf{2.34} & \textbf{2.18} & \textbf{2.21} & \textbf{2.88} & \textbf{2.87} & \textbf{1.02} & \textbf{2.25}
    \end{tabular}
    }
    
    
    \caption{Averages of the values presented in Figure~\ref{fig:sd-analysis-tda-tree-permutations} (number of standard deviations away from the mean). In each block we fix one of the parameters, and aggregate over all others.}
    \label{tab:aggregated_results_std}
\end{table}

\begin{table}[htb]
    \centering
    \scriptsize
    {
    \setlength{\tabcolsep}{4pt}
    \begin{tabular}{l|llllll|l}
        \# $\geq$ 99500 & j\_k1      & j\_k2     & mat       & phy       & clust       & path       & sum         \\[.8ex]
        \hline &&&&&&& \\[-2ex]
        Euclidean, dim 0   & 6           & \textbf{6} & 2           & 6           & 4           & 0          & 24          \\
        Euclidean, dim 1   & \textbf{8}  & 4          & 4           & 7           & 12          & 0          & 35          \\
        Euclidean, dim 2   & 7           & 3          & \textbf{8}  & 6           & 9           & 0          & 33          \\
        cosine, dim 0      & 2           & 1          & 2           & 5           & 7           & 0          & 17          \\
        cosine, dim 1      & 5           & 3          & 6           & \textbf{13} & \textbf{14} & \textbf{1} & \textbf{42} \\
        cosine, dim 2      & 2           & 0          & 4           & 10          & 10          & 0          & 26          \\[.8ex]
        \hline &&&&&&& \\[-2ex]
        bottleneck         & 4           & 2          & 3           & 9           & 12          & 0          & 30          \\
        sliced Wasserstein & 10          & \textbf{7} & 9           & 14          & \textbf{19} & 0          & 59          \\
        persistence image  & 5           & 3          & 4           & 7           & 8           & 0          & 27          \\
        bars statistics    & \textbf{11} & 5          & \textbf{10} & \textbf{17} & 17          & \textbf{1} & \textbf{61} \\[.8ex]
        \hline &&&&&&& \\[-2ex]
        nj                 & 12          & 7           & 8           & 15          & 21          & 0          & 63           \\
        upgma              & \textbf{18} & \textbf{10} & \textbf{18} & \textbf{32} & \textbf{35} & \textbf{1} & \textbf{114}
    \end{tabular}
    }
    \caption{The number of dots in Figure~\ref{fig:sd-analysis-tda-tree-permutations} corresponding to parameters that performed better than 99,500 random permutations of the tree leaves (labeled by star in the figure). The total number of dots for a fixed distance that meet the parameters in each block is 24, 36 and 72, respectively.}
    \label{tab:aggregated_results_geq99500}
\end{table}

\begin{table}[htb]
    \centering
    \scriptsize{
    \setlength{\tabcolsep}{4pt}
    \begin{tabular}{l|rrrrrrr}
     \makebox[0pt][l]{full fig}\phantom{30 lang}  &   all &   jrf\_k1 &   jrf\_k2 &   match &   phylo &   clust &   path \\[.2ex]
    \hline &&&&&&& \\[-2ex]
     total               &   864 &      144 &      144 &     144 &     144 &     144 &    144 \\[.5ex]
     $\sigma > 1$     &   681 &      121 &      111 &     117 &     134 &     134 &     64 \\
     $\sigma > 2$     &   451 &       84 &       74 &      72 &      99 &     103 &     19 \\
     $\sigma > 3$     &   229 &       36 &       41 &      29 &      59 &      63 &      1 \\
     $\sigma > 4$     &    74 &       13 &       15 &       7 &      17 &      22 &      0 \\
     $\sigma > 5$     &    17 &        2 &        4 &       1 &       5 &       5 &      0 \\
     $\sigma > 6$     &     5 &        0 &        2 &       0 &       2 &       1 &      0
    \end{tabular}
    \vspace{1ex}
    
    \begin{tabular}{l|rrrrrrr}
     30 lang   &   all &   jrf\_k1 &   jrf\_k2 &   match &   phylo &   clust &   path \\[.2ex]
    \hline &&&&&&& \\[-2ex]
     total               &   288 &       48 &       48 &      48 &      48 &      48 &     48 \\[.5ex]
     $\sigma > 1$     &   242 &       43 &       40 &      43 &      44 &      45 &     27 \\
     $\sigma > 2$     &   155 &       31 &       31 &      24 &      26 &      30 &     13 \\
     $\sigma > 3$     &    69 &       12 &       19 &      11 &      13 &      13 &      1 \\
     $\sigma > 4$     &    22 &        3 &        4 &       3 &       4 &       8 &      0 \\
     $\sigma > 5$     &     7 &        1 &        1 &       1 &       2 &       2 &      0 \\
     $\sigma > 6$     &     4 &        0 &        1 &       0 &       2 &       1 &      0
    \end{tabular}
    \vspace{1ex}
    
    \begin{tabular}{l|rrrrrrr}
     50 lang   &   all &   jrf\_k1 &   jrf\_k2 &   match &   phylo &   clust &   path \\[.2ex]
    \hline &&&&&&& \\[-2ex]
     total               &   288 &       48 &       48 &      48 &      48 &      48 &     48 \\[.5ex]
     $\sigma > 1$     &   230 &       40 &       36 &      39 &      47 &      46 &     22 \\
     $\sigma > 2$     &   162 &       28 &       26 &      24 &      40 &      40 &      4 \\
     $\sigma > 3$     &    92 &       16 &       17 &      10 &      23 &      26 &      0 \\
     $\sigma > 4$     &    39 &        8 &        9 &       2 &      10 &      10 &      0 \\
     $\sigma > 5$     &     9 &        1 &        3 &       0 &       3 &       2 &      0 \\
     $\sigma > 6$     &     1 &        0 &        1 &       0 &       0 &       0 &      0
    \end{tabular}
    \vspace{1ex}
    
    \begin{tabular}{l|rrrrrrr}
     81 lang   &   all &   jrf\_k1 &   jrf\_k2 &   match &   phylo &   clust &   path \\[.2ex]
    \hline &&&&&&& \\[-2ex]
     total               &   288 &       48 &       48 &      48 &      48 &      48 &     48 \\[.5ex]
     $\sigma > 1$     &   209 &       38 &       35 &      35 &      43 &      43 &     15 \\
     $\sigma > 2$     &   134 &       25 &       17 &      24 &      33 &      33 &      2 \\
     $\sigma > 3$     &    68 &        8 &        5 &       8 &      23 &      24 &      0 \\
     $\sigma > 4$     &    13 &        2 &        2 &       2 &       3 &       4 &      0 \\
     $\sigma > 5$     &     1 &        0 &        0 &       0 &       0 &       1 &      0 \\
     $\sigma > 6$     &     0 &        0 &        0 &       0 &       0 &       0 &      0
    \end{tabular}
    }
    \caption{The number of parameters in Figure~\ref{fig:sd-analysis-tda-tree-permutations} with value greater than a given threshold.}
    \label{tab:dots_greater_than_sigma}
\end{table}

\begin{table}[htb]
    \centering
    \scriptsize{
    \setlength{\tabcolsep}{4pt}
    \begin{tabular}{l|rrrrrrr}
     full fig   &   all &   jrf\_k1 &   jrf\_k2 &   match &   phylo &   clust &   path \\[.2ex]
    \hline &&&&&&& \\[-2ex]
     total            &   864 &      144 &      144 &     144 &     144 &     144 &    144 \\[.5ex]
     top $10.00\%$ &   613 &      113 &      102 &      98 &     121 &     128 &     51 \\
     top $\hphantom{1}5.00\%$ &   484 &       90 &       75 &      76 &     102 &     112 &     29 \\
     top $\hphantom{1}1.00\%$ &   255 &       39 &       34 &      42 &      65 &      72 &      3 \\
     top $\hphantom{1}0.50\%$ &   177 &       30 &       17 &      26 &      47 &      56 &      1 \\
     top $\hphantom{1}0.10\%$ &    57 &       10 &        4 &      10 &      14 &      19 &      0 \\
     top $\hphantom{1}0.05\%$ &    32 &        5 &        3 &       5 &       9 &      10 &      0 \\
     top $\hphantom{1}0.01\%$ &    10 &        1 &        2 &       2 &       2 &       3 &      0 \\
    \end{tabular}
    \vspace{1ex}
    
    \begin{tabular}{l|rrrrrrr}
     30 lang    &   all &   jrf\_k1 &   jrf\_k2 &   match &   phylo &   clust &   path \\[.2ex]
    \hline &&&&&&& \\[-2ex]
     total            &   288 &       48 &       48 &      48 &      48 &      48 &     48 \\[.5ex]
     top $10.00\%$ &   217 &       40 &       39 &      33 &      36 &      44 &     25 \\
     top $\hphantom{1}5.00\%$ &   172 &       34 &       33 &      26 &      28 &      34 &     17 \\
     top $\hphantom{1}1.00\%$ &    75 &       13 &       15 &      14 &      12 &      18 &      3 \\
     top $\hphantom{1}0.50\%$ &    46 &       11 &        5 &       9 &       9 &      11 &      1 \\
     top $\hphantom{1}0.10\%$ &    16 &        2 &        1 &       3 &       4 &       6 &      0 \\
     top $\hphantom{1}0.05\%$ &     9 &        1 &        1 &       3 &       2 &       2 &      0 \\
     top $\hphantom{1}0.01\%$ &     5 &        1 &        1 &       1 &       1 &       1 &      0 \\
    \end{tabular}
    \vspace{1ex}
    
    \begin{tabular}{l|rrrrrrr}
     50 lang    &   all &   jrf\_k1 &   jrf\_k2 &   match &   phylo &   clust &   path \\[.2ex]
    \hline &&&&&&& \\[-2ex]
     total            &   288 &       48 &       48 &      48 &      48 &      48 &     48 \\[.5ex]
     top $10.00\%$ &   212 &       38 &       33 &      34 &      47 &      43 &     17 \\
     top $\hphantom{1}5.00\%$ &   169 &       30 &       25 &      25 &      41 &      41 &      7 \\
     top $\hphantom{1}1.00\%$ &   100 &       17 &       15 &      13 &      27 &      28 &      0 \\
     top $\hphantom{1}0.50\%$ &    74 &       13 &       10 &       9 &      18 &      24 &      0 \\
     top $\hphantom{1}0.10\%$ &    29 &        6 &        3 &       3 &       8 &       9 &      0 \\
     top $\hphantom{1}0.05\%$ &    18 &        3 &        2 &       1 &       6 &       6 &      0 \\
     top $\hphantom{1}0.01\%$ &     3 &        0 &        1 &       0 &       1 &       1 &      0 \\
    \end{tabular}
    \vspace{1ex}
    
    \begin{tabular}{l|rrrrrrr}
     81 lang    &   all &   jrf\_k1 &   jrf\_k2 &   match &   phylo &   clust &   path \\[.2ex]
    \hline &&&&&&& \\[-2ex]
     total            &   288 &       48 &       48 &      48 &      48 &      48 &     48 \\[.5ex]
     top $10.00\%$ &   184 &       35 &       30 &      31 &      38 &      41 &      9 \\
     top $\hphantom{1}5.00\%$ &   143 &       26 &       17 &      25 &      33 &      37 &      5 \\
     top $\hphantom{1}1.00\%$ &    80 &        9 &        4 &      15 &      26 &      26 &      0 \\
     top $\hphantom{1}0.50\%$ &    57 &        6 &        2 &       8 &      20 &      21 &      0 \\
     top $\hphantom{1}0.10\%$ &    12 &        2 &        0 &       4 &       2 &       4 &      0 \\
     top $\hphantom{1}0.05\%$ &     5 &        1 &        0 &       1 &       1 &       2 &      0 \\
     top $\hphantom{1}0.01\%$ &     2 &        0 &        0 &       1 &       0 &       1 &      0 \\
    \end{tabular}
    }
    \caption{The number of parameters in Figure~\ref{fig:sd-analysis-tda-tree-permutations} with distance smaller than the given percentage of the 100,000 permutations.}
    \label{tab:dots_in_top_percentage}
\end{table}

\newpage

\begin{figure*}[ht]
    \centering
    \includegraphics[width=1\linewidth]{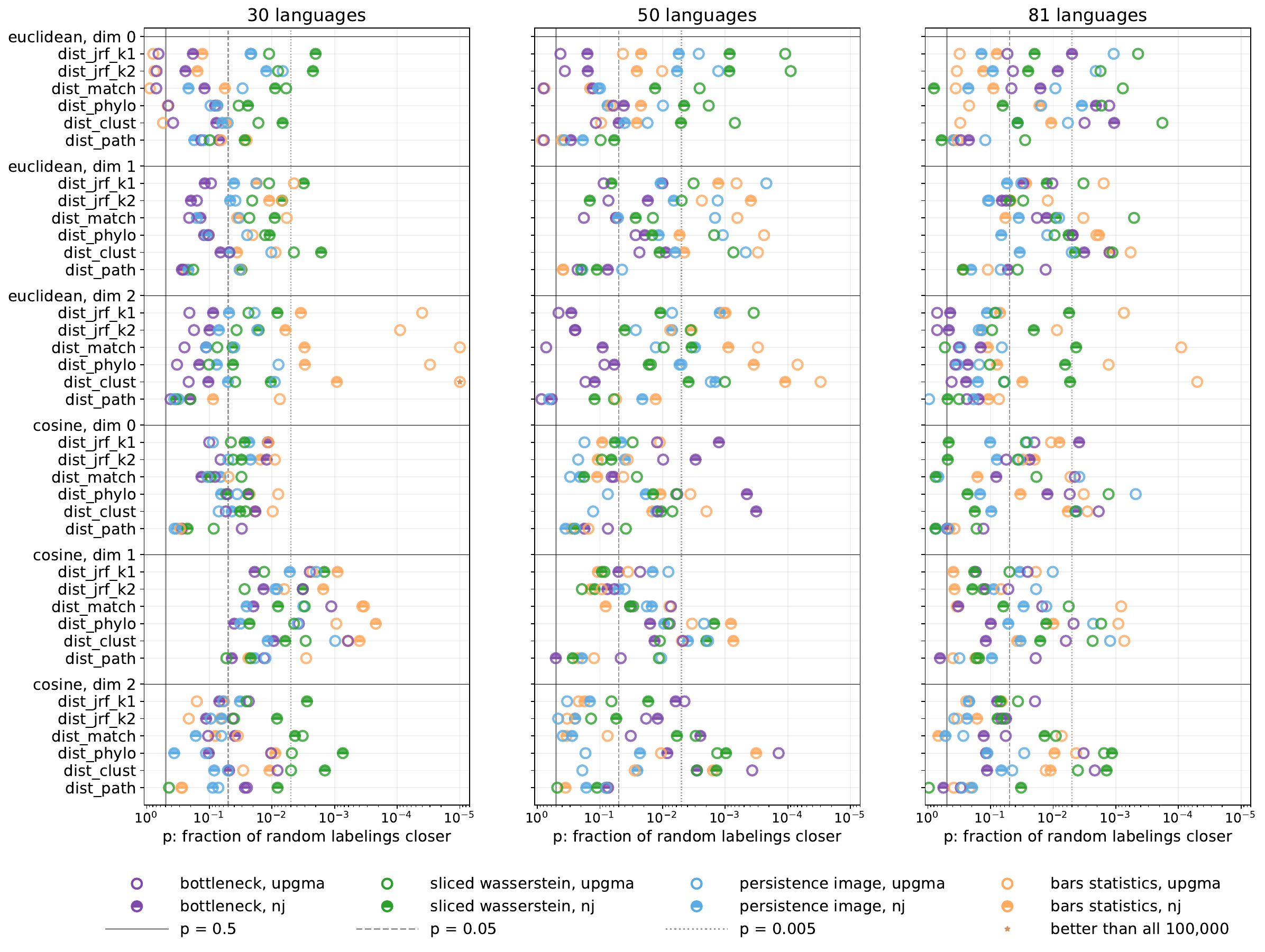}
    \caption{The statistical significance of TDA trees for 30, 50, and 81 languages against the Ethnologue reference for trees reconstructed by UPGMA and NJ for each combination of parameters described in Section~\ref{sec:methods}, as in Figure~\ref{fig:sd-analysis-tda-tree-permutations}. Each dot represents the distance of a single reconstructed tree to the reference, and its position shows what fraction of 100,000 random permutations of the tree's leaves lead to smaller distances.}
    \label{fig:sd-analysis-tda-tree-permutations_p-values}
\end{figure*}

\begin{figure*}
    \centering
    \begin{subfigure}{.3\textwidth}
        \centering
        \includegraphics[height=87mm]{figs/ethnologue_tree_30.pdf}
        \caption{Reference Ethnologue tree}
    \end{subfigure}\hfill
    \begin{subfigure}{.35\textwidth}
        \centering
        \includegraphics[height=87mm]{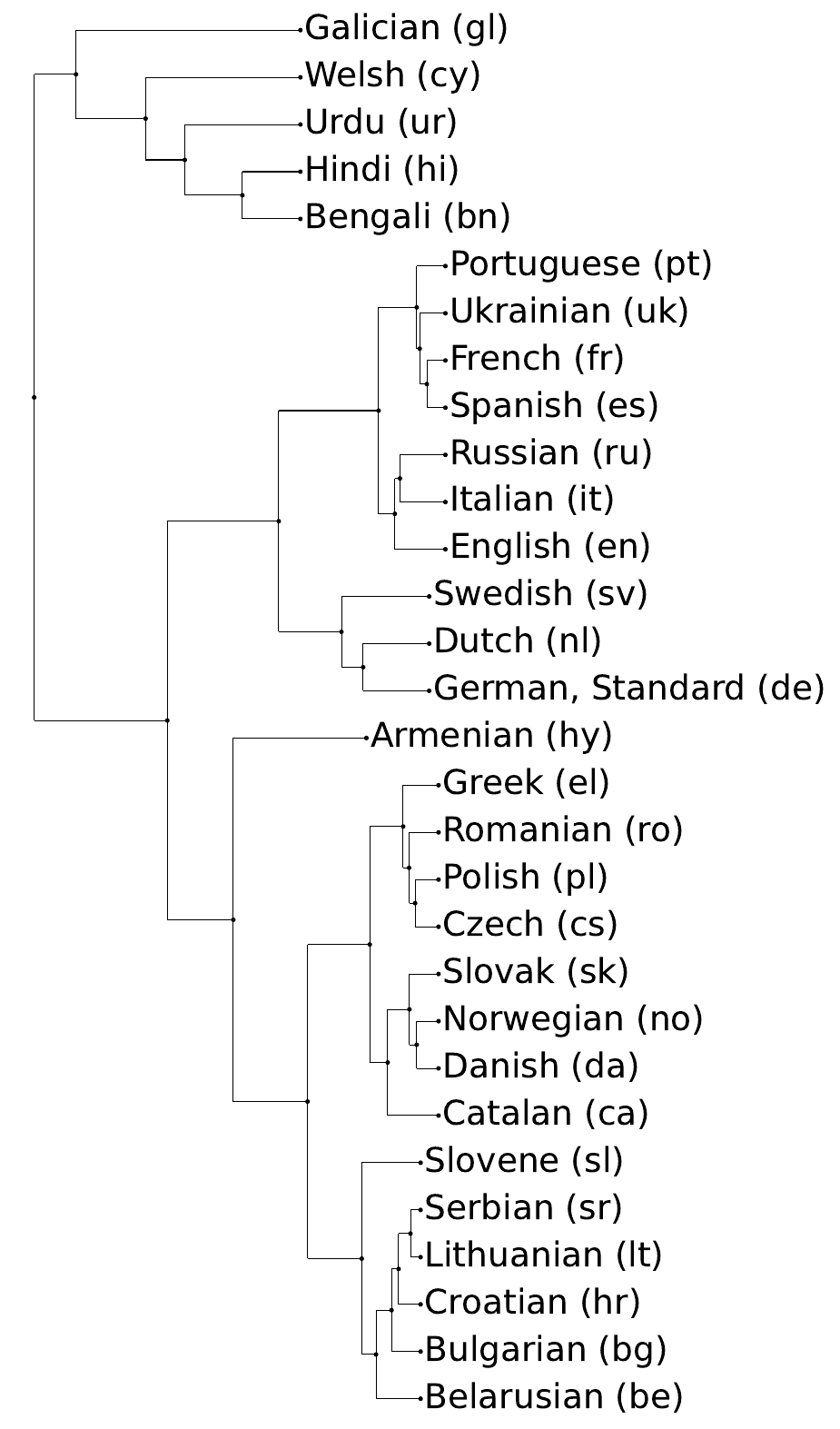}
        \caption{UPGMA}
    \end{subfigure}\hfill
    \begin{subfigure}{.3\textwidth}
        \centering
        \includegraphics[height=87mm]{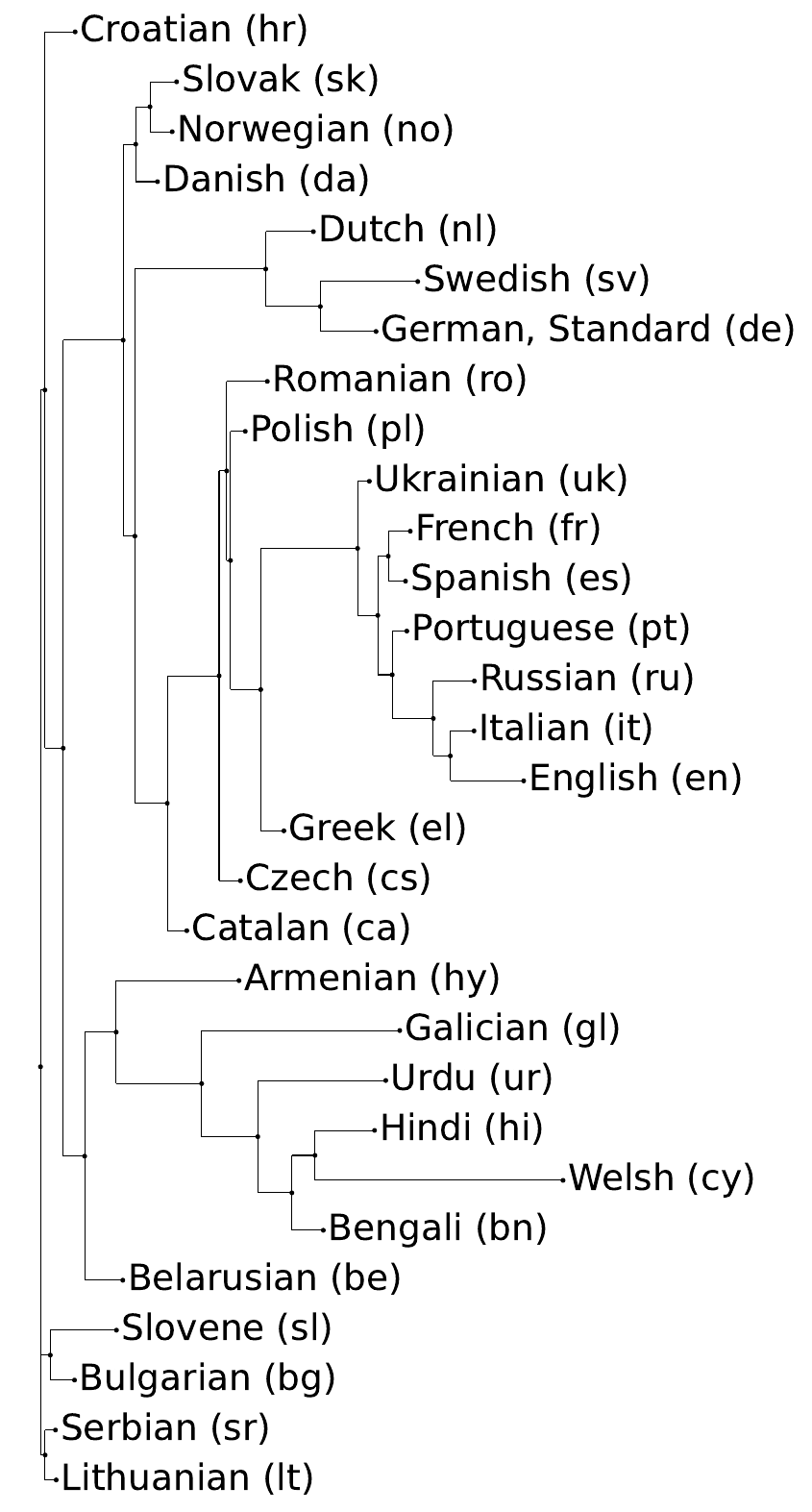}
        \caption{NJ}
    \end{subfigure}
    \caption{The 30-language reference Ethnologue tree with representative UPGMA and NJ reconstructed trees using TDA (here 2-dimensional persistent homology using Euclidean distance and inferred distances between languages based on bars statistics between the persistent diagrams).
    Although far from perfect agreement, the reconstructed trees captures many meaningful language subgroups; for UPGMA, e.g. (i) clusters of Balto-Slavic languages (\emph{sl}, \emph{sr}, \emph{lt}, \emph{hr}, \emph{bg}, \emph{be}), 
    (ii) the Romance Italo-Western group (\emph{pt}, \emph{fr}, \emph{es}, \emph{it}) mixed with Germanic languages (\emph{de}, \emph{nl}, \emph{en}, \emph{sv}), while
    (iii) the more distant Indo-Aryan languages (\emph{bn}, \emph{hi}, \emph{ur}) are clustered and well-separated.}
    
    \label{fig:upgma_nj_trees}
\end{figure*}

\end{document}